%% file: diff_length_tsc.tex
\documentclass[smallcondensed]{svjour3}

\input{preamble}

\begin{document}
	\title{Time series classification for varying length series}
	\author{Chang Wei Tan \and Fran\c{c}ois Petitjean \and Eamonn Keogh \and Geoffrey I. Webb}
	\institute{Chang Wei Tan \and Fran\c{c}ois Petitjean \and Geoffrey I. Webb \at Faculty of Information Technology\\
	25 Exhibition Walk\\
	Monash University, Melbourne\\
	VIC 3800, Australia\\
	\email{chang.tan@monash.edu,francois.petitjean@monash.edu,geoff.webb@monash.edu}\\
	Eamonn Keogh \at 
    Computer Science \& Engineering Department\\
    University of California - Riverside\\
    Riverside, CA 92521\\
    \email{eamonn@cs.ucr.edu}
    }
        
	\maketitle
	
	\begin{abstract}
		Research into time series classification has tended to focus on the case of series of uniform length.
		However, it is common for real-world time series data to have unequal lengths.
		Differing time series lengths may arise from a number of fundamentally different mechanisms. 
		In this work, we identify and evaluate two classes of such mechanisms -- variations in sampling rate relative to the relevant signal and variations between the start and end points of one time series relative to one another.
		We investigate how time series generated by each of these classes of mechanism are best addressed for time series classification.
		We perform extensive experiments and provide practical recommendations on how variations in length should be handled in time series classification. 
	\end{abstract}
	
	\section{Introduction}
	\label{sec:introduction}
	
	Time series classification (TSC) is an important task in many modern world applications such as remote sensing \citep{pelletier2019temporal,petitjean2012satellite},
	astronomy \citep{batista2011complexity},
	speech recognition \citep{hamooni2016phoneme}
	, and insect classification \citep{chen2014flying}.
	The time series to be classified are the observed outputs generated by some process. The classification task often relates to identifying the class of the process that generated the series. 
	
	Each class of process might be considered as a realization of one or more ideals (in the Platonic sense) or prototypes.
	The resulting time series can then be conceived as corrupted variants of ideal or prototypical time series.
	An observed time series might differ from the ideal in many ways.
	Much of the research on time series distance measures in the last decade can be seen as the introduction of techniques to mitigate these differences, either as a preprocessing step or directly in a distance measure.
	For example, variations in amplitude and offset are typically addressed in time series classification by normalization of the series \citep{rakthanmanon2012searching}. 
	Some observed values may be erroneous and might be addressed by outlier detection \citep{basu2007automatic} and subsequent reinterpolation \citep{pelletier2019temporal}. 
	Local variations of timing are typically addressed by Dynamic Time Warping (DTW) \citep{rakthanmanon2012searching}.
	Finally, out of phase time series subsequences are typically compared using a phase invariant measure \citep{batista2011complexity}.
	
	In this paper we address the issue of time series that vary from the ideal form in that they differ in length. 
	In particular, we posit that there are a few fundamental types of mechanism by which the time series generated by variants of a single prototypical process may end up having different lengths. 
	At first blush, the problem of handling different length time series may seem trivial. To see that this is not the case, let us consider the analogous case where we must compare discrete strings using the Hamming distance, a natural proxy for the Euclidean distance. Suppose we encounter the following pair of strings of unequal length: A = bat, B = batman.
    It is clear that truncation of the longer string is the correct thing to do here. However, what about this pair of strings? C = bat, D = bbaatt.
    It is equally obvious that truncation would be a mistake here. 
    In this case we need to “stretch out” the shorter word.
    As we discuss below, both this situations have analogous cases in real-valued time series.

	One of these mechanisms corresponds to variation in the relative frequency at which the process is observed. For instance, the generating processes might unfold at differing speeds, or the sensors might operate at differing frequencies. 
	By way of example, in remote sensing applications, the crops being observed may grow at varying rates due to different climatic conditions (eg sun, water) from one crop to another \citep{petitjean2012satellite}. 
	The dual to this variation of speed of the observed phenomenon is when the observation itself is irregular or operating at varying speeds. This is for example the case, again in remote sensing, where observations might be missing at certain dates because of the presence of clouds. This is also the case when a wrist-worn heart-rate monitor decides to decrease the sensing frequency when the person doesn't move much (i.e.~is likely to have a stable heart rate), and increase it when the person becomes active. These two types of variations (speed of the phenomenon or frequency of observation) are duals of one another and treated as one type of generation in this study. 
	
	Another mechanism is variation with respect to the points during the process at which the recorded observations begin and end. For example, the start and end of an audio recording might be decided upon manually, resulting in signals that have different lengths. The points at which, say, one phoneme begins and another ends, during recorded speech, might not be clear and so there may be some randomness in when an extracted series begins and ends.
	
	A handful of strategies have been devised to address classification of time series with differing lengths.
	However, compared to the “time warping” problem about which there are hundreds of research papers, this issue seems to be relatively understudied and unappreciated. We suspect that this is simply because the UCR Archive has long had dozens of examples of the former, and none of the latter, although the newest iteration of this resources has attempted to rectify this \citep{UCRArchive2018}.
	
\begin{figure}
		\centering
		\begin{subfigure}[]{0.49\linewidth}
			\centering
			\includegraphics[width=\linewidth]{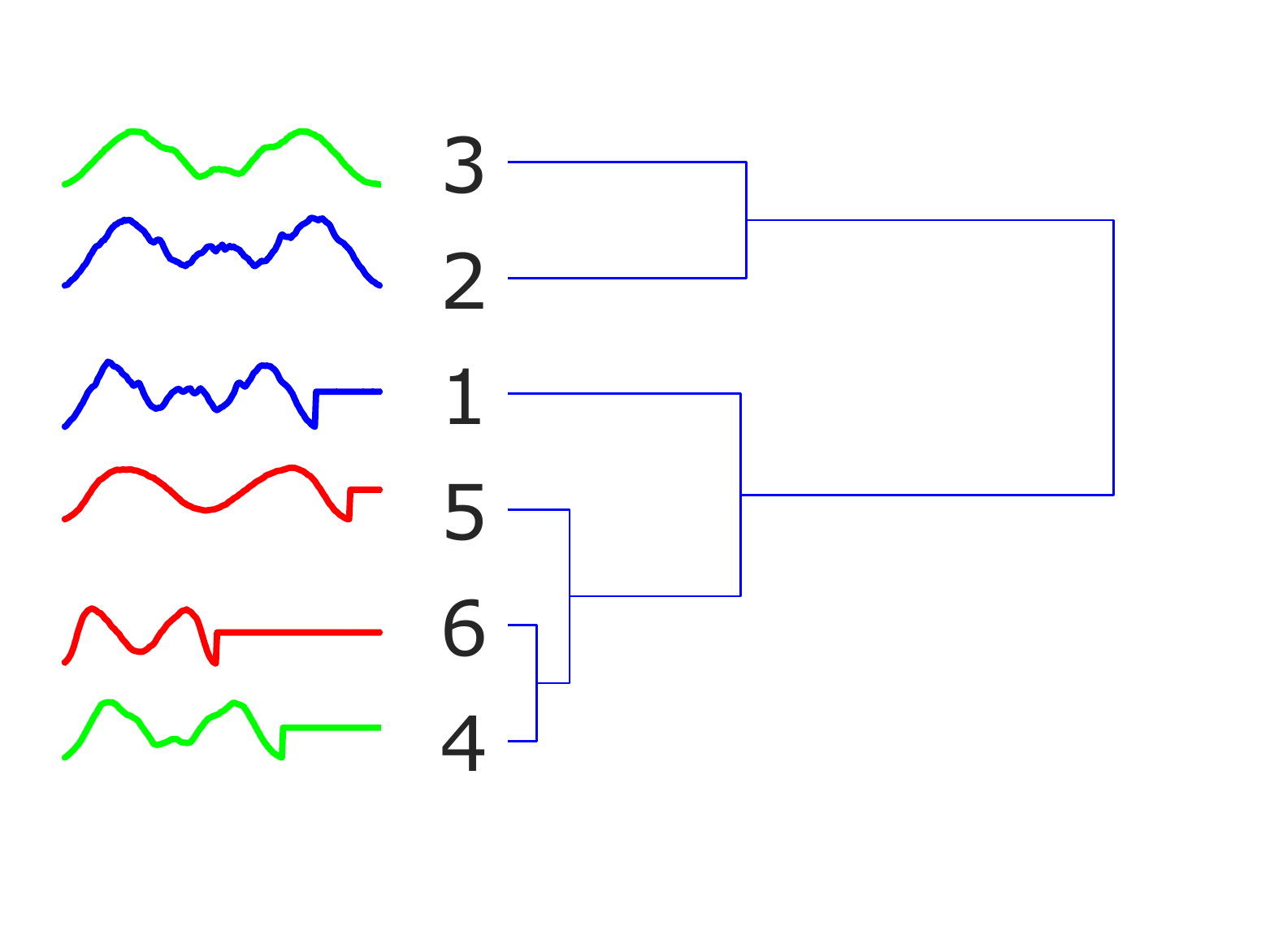}
			\vspace*{-28pt}\caption{}
			\label{fig:suffix noise}
		\end{subfigure}
		\hfill
		\begin{subfigure}[]{0.49\linewidth}
			\centering
			\includegraphics[width=\linewidth]{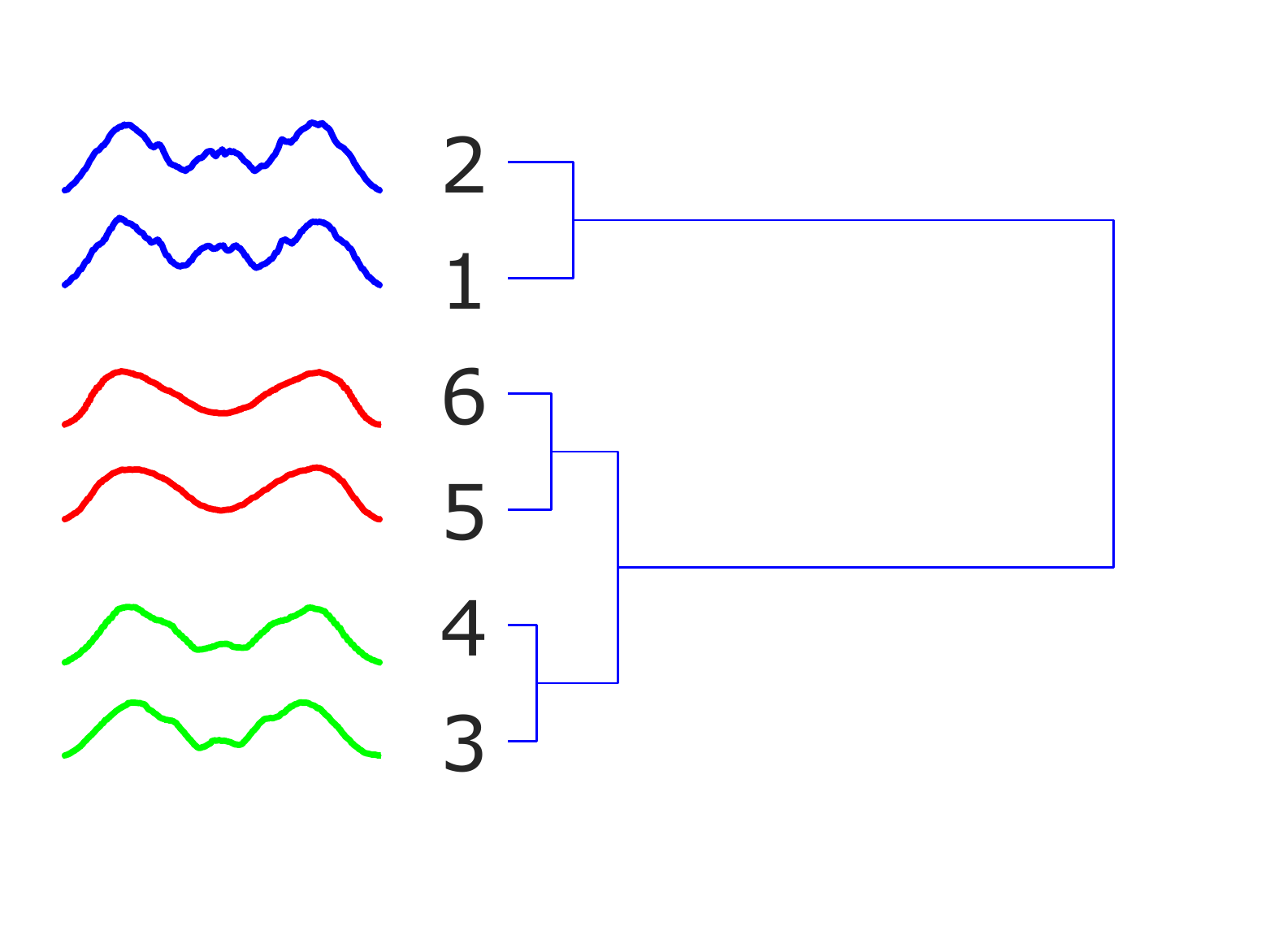}
			\vspace*{-28pt}\caption{}
			\label{fig:rescale}
		\end{subfigure}
		\hfill
		\begin{subfigure}[]{0.49\linewidth}
			\centering
			\includegraphics[width=\linewidth]{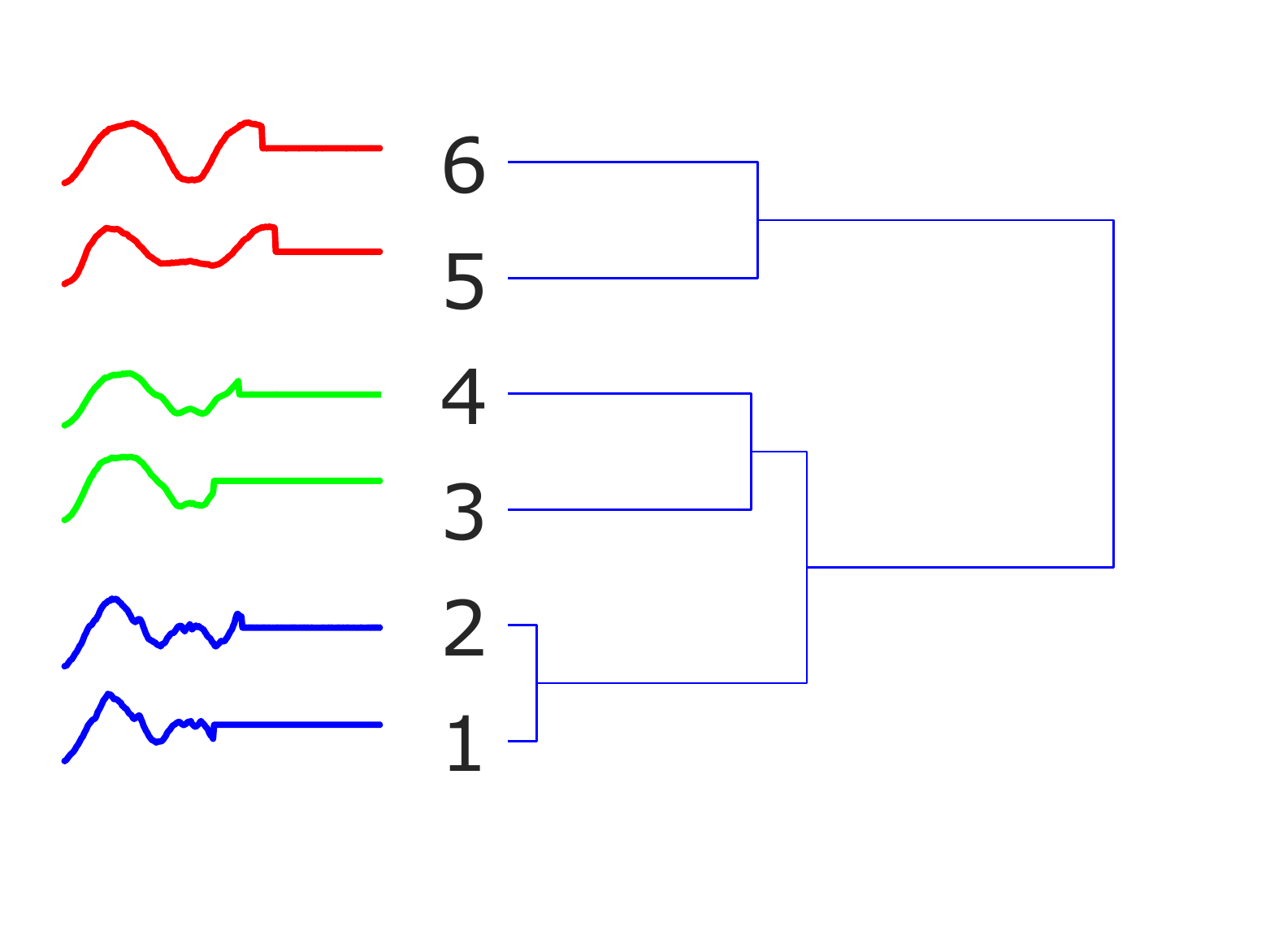}
			\vspace*{-28pt}\caption{}
			\label{fig:prefix with suffix noise}
		\end{subfigure}
		\hfill
		\begin{subfigure}[]{0.49\linewidth}
			\centering
			\includegraphics[width=\linewidth]{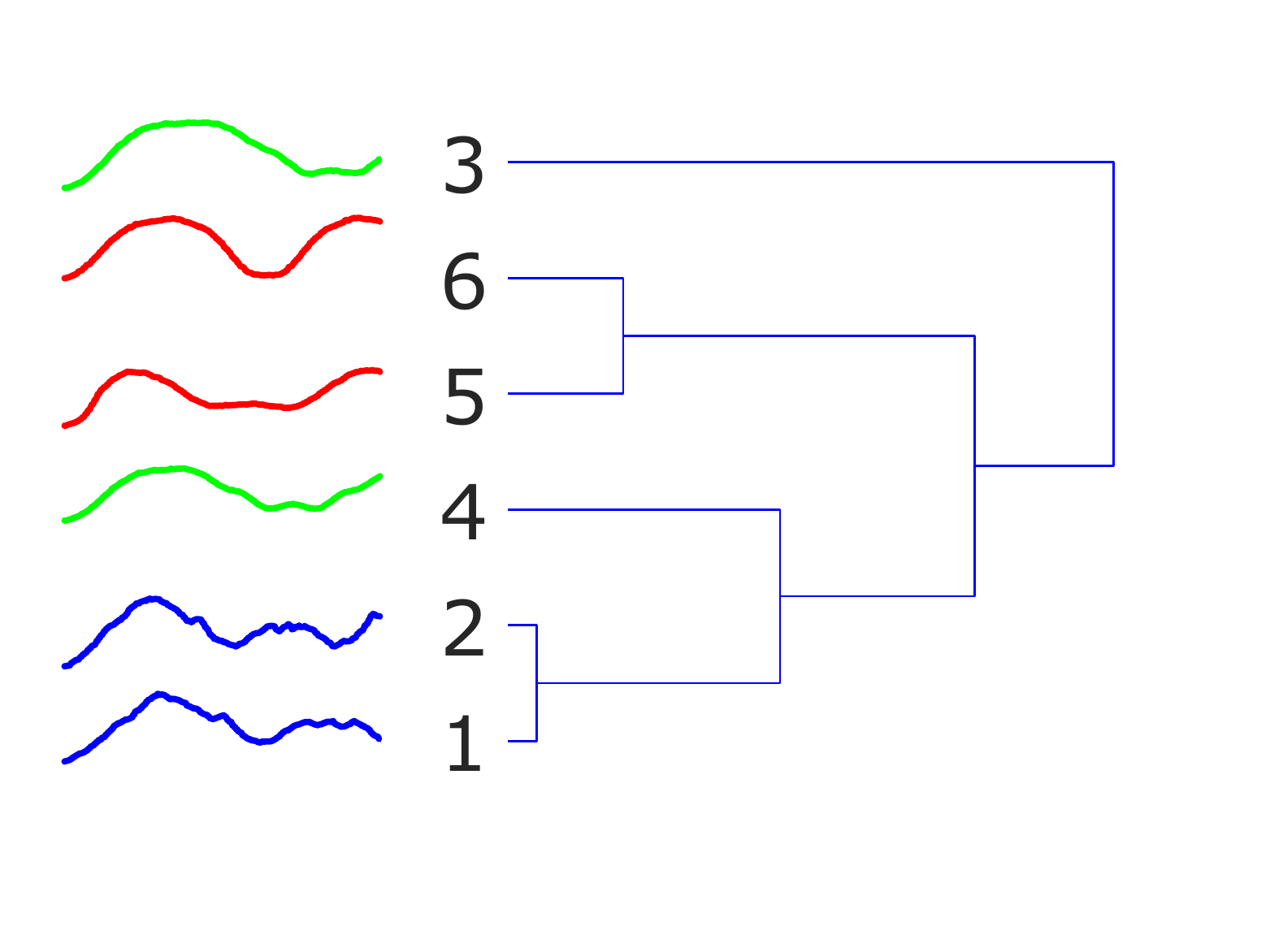}
			\vspace*{-28pt}\caption{}
			\label{fig:prefix with rescale}
		\end{subfigure}
		\caption{Dendrograms, computed with \dtw{} distance, of six series, two from each of three classes, with differing length due to differing sampling rates (a) padded with low amplitude random noise and (b) rescaled to the same length. When the series differ in length due to differing sampling rates, rescaling recovers the correct classes. Dendrograms where series differ in length due to differing end points (c) padded with low amplitude random noise and (d) rescaled to the same length. When the series differ in length due to differing end points, padding recovers the correct classes.}
		\label{fig:dendrograms}
	\end{figure}

	This paper systematically explores how existing strategies perform with differing mechanisms for affecting series length. It further explores how these strategies and mechanisms interact with different types of time series classifier. 
	As an example, if the time series differ in length due to differing sampling rates,
	uniform rescaling to equal length is more intuitive than adding noise to the suffix. 
	This is illustrated in Figures \ref{fig:suffix noise} and \ref{fig:rescale} where the time series from the \texttt{ArrowHead} dataset are resized using differing sampling rates. %
	We expect that time series in the same class (the same color in the Figures) will be grouped together under the dynamic time warping (\dtw{}) distance. 
	However, Figure \ref{fig:suffix noise} shows that when low amplitude noise are added to the suffix, series 3 and 4 are separated despite belonging to the same class.
	With rescaling, the correct grouping is achieved (Figure \ref{fig:rescale}). One might then think that rescaling all the series to the same length is always a good solution. We show that this is not true if, for instance, the process that generated the data with different lengths was one of stopping the recording earlier or later. In that case, we see in Figure \ref{fig:prefix with suffix noise} that the correct grouping is obtained by padding the suffix with zeros, while uniformly rescaling (Figure \ref{fig:prefix with rescale}) leads to the wrong grouping --~with series 3 and 4 again being separated.
	
	Our paper is organized as follows. 
	In Section \ref{sec:tsc}, we introduce some key TSC algorithms.
	Then we describe in Section \ref{sec:generators} some of the mechanisms that result in series lengths differing. In Section \ref{sec:processors} we discuss the key processing techniques that have been proposed to preprocess a collection of varying length series to be of uniform length.
	In Section \ref{sec:related work}, we review some related work on handling time series of variable lengths.
	We evaluate the interactions between length variation mechanisms, preprocessing methods and classification algorithms in Section \ref{sec:experiment}. We~conclude with future work in Section \ref{sec:conclusion}.
	
	\section{Time series classification}
	\label{sec:tsc}
	Numerous time series classification (TSC) algorithms have been proposed \citep{bagnall2017great,fawaz2018deep},
	from the classic distance-based nearest neighbour (\nn{}) approaches \citep{keogh2005exact,chen2004marriage,tan2017indexing,wang2013experimental}, bag-of-word approaches \citep{senin2013sax,schafer2015boss}, time series shapelets approaches \citep{ye2009time,mueen2011logical,hills2014classification,grabocka2016fast,rakthanmanon2013fast}, ensemble-based approaches \citep{schafer2015boss,lines2015time,schafer2017fast,lucas2019proximity}, to deep learning approaches \citep{fawaz2018deep,wang2017time}.
	In this section, we briefly discuss some key existing TSC algorithms, with an emphasis on the main overarching strategies rather than on providing a detailed review such as can be found in \citep{bagnall2017great}. 
	
	\subsection{Classic distance-based TSC strategies}
	\label{subsec:nn}
	For more than a decade, the distance-based \nn{} approaches have been the algorithms of choice for TSC.
	The \nn{} algorithm is typically coupled with a distance measure to compare a pair of time series.
	The paper \citep{lines2015time} summarizes the top distance measures used with the \nn{} classifier for TSC.
	Among all these distance measures, the most competitive and popular are the Euclidean distance and \dtw{} \citep{ding2008querying}.
	The Euclidean distance is known for its simplicity and good classification accuracy, especially when the training dataset is large, as shown in \citep{ding2008querying}.

	On the other hand, the \dtw{} distance is more computational expensive, but subsumes Euclidean distance. Thus, with appropriate parametrization, \dtw{} can describe a wider range of relevant similarity functions. 
	In fact, \nndtw{} with its warping window set through cross validation was a strong baseline for TSC algorithms for more than a decade \citep{wang2013experimental,lines2015time,bagnall2017great}.
	We refer interested readers to the paper \citep{lines2015time} for the technical details of the different distance measures.
	Although \dtw{} can be used to compare unequal length time series \citep{ratanamahatana2005three}, the warping window does not address this well.
	For example, the result is undefined if the warping window is smaller than the difference in length between the series, leaving no possible path joining the 2 corner points of the warping matrix.
	
	
	Other than Euclidean distance and \dtw{}, the \nn{} classifier can also be coupled with other distance measures \citep{lines2015time}, each tackling time series from different domains. 
	The Ensemble of Elastic Distances (\ee{}) \citep{lines2015time} combines eleven \nn{} classifiers, each using a different distance measure.
	It achieves lower error than any one of the constituent classifiers on the majority of UCR benchmark datasets.
	\ee{}, however, has high complexity which prohibits it from being practical.
	The Proximity Forest (\pf{}) \citep{lucas2019proximity} was proposed as a close relative to \ee{}.
	It is an ensemble of proximity trees where each tree is a combination of multiple distance measures. 
	The training process first involves choosing a random time series per class and compare them to every instance in the training set using a random distance measure.
	The training set is then split based on the proximity of the training set to the exemplars and the process repeated until the leaf nodes are pure.
	\pf{} has been shown to be more accurate and significantly more scalable than \ee{} \citep{lucas2019proximity}.
	
	\subsection{Time series shapelets algorithms}
	\label{subsec:shapelets}
	In addition to comparing the distances between two time series, there are also algorithms that extract features from the time series, such as time series shapelets. 
	A shapelet is a subsequence of a time series that gives the optimal split for the training dataset and can be used to discriminate between the different classes by building a decision tree \citep{ye2009time}.
	A problem with this algorithm is the high number of shapelet candidates to test. 
	This led to the development of faster shapelets algorithms, such as Fast Shapelets \citep{rakthanmanon2013fast}, which approximates the shapelets using Symbolic Aggregation Approximation (SAX). 
	The Learned Shapelets algorithm \citep{grabocka2014learning} uses K-means clustering to speed up the shapelet discovery process. 
    The more recent Shapelet Transform algorithm \citep{hills2014classification} transforms the time series using the distance of the time series to all shapelet candidates, which act as the features of the time series. 
    Then these features are used to construct the Shapelet Ensemble algorithm, the second most accurate time series classifier in the great time series classification bakeoff \citep{bagnall2017great}. 
	
	\subsection{Bag-of-words algorithms}
	\label{subsec:bag of words}
	Unlike distance-based approaches,
	the bag-of-words algorithms measure the similarity between two time series using the distribution of words.
	A word is a discretised version of a subsequence in the time series.
	Typically a bag-of-words algorithm slides a window across the time series to extract all subsequences, label each of the subsequence with symbols and create a histogram per time series.
	The Bag-of-Pattern (BOP) \citep{lin2012rotation} algorithm labels the subsequence using SAX words. 
	The SAX-VSM algorithm \citep{senin2013sax} combines the original BOP algorithm with vector-space model which is more accurate and scalable than BOP.
	The \boss{} Ensemble \citep{schafer2015boss} transforms the time series into Symbolic Fourier Approximation (SFA) words.
	The subsequences are transformed into SFA using truncated Discrete Fourier Transform (DFT) and discretised using Multiple Coefficient Binning (MCB) technique \citep{schafer2015boss}.
	Among the three algorithms, \boss{} is the most accurate and has significantly better rank on accuracy across the UCR datasets than \nndtw{} \citep{bagnall2017great}. 
	It has worse rank than \pf{}, but not significantly worse.
	
	\subsection{Ensemble-based algorithms} 
	\label{subsec:ensembles}
	A new generation of ensemble classifiers have recently established new benchmarks in time series classification accuracy.
	\ee{} \citep{lines2015time} was one of the first classifiers that was more accurate on the UCR datasets than the previous benchmark \nndtw{}. 
	It has been followed by \pf{}, \boss{}, Shapelet Ensemble and the Collection of Transformation-based Ensembles (\cote{}) \citep{bagnall2015time}.
	\cote{} is a meta-ensemble that consists of 35 classifiers in different transformation domains, including 11 from \ee{} and 8 from Shapelet Ensemble. 
	The limitation of \cote{} is that it is more biased towards the domain with more classifiers as each of the classifier is treated as a single module. 
	This limitation is overcome by the Hierarchical-Vote Collection of Transformation-based Ensemble (\hivecote{}) \citep{lines2016hive} --- a meta-ensemble consisting of multiple other ensembles including \ee{}, \cote{}, \boss{} and Shapelet Ensemble.
	Each of these ensembles is treated as a single module and voting is adjusted to prevent bias towards modules containing more classifiers.
	A major study established that \hivecote{} attains the best rank on classification accuracy of all major TSCs on the UCR datasets \citep{lines2016hive}.
	
	Most ensemble-based algorithms have extremely high computational complexity. The distance-based \pf{} algorithm achieves competitive accuracy, with orders of magnitude less computation, by combining the efficiency of tree-based classifiers with the accuracy of specialized time-series distance measures \citep{lucas2019proximity}.
	
	\section{Varying lengths time series}
	\label{sec:generators}
	Time series for a single classification task may vary in length for a wide range of reasons. 
	In this section, we describe two fundamental classes of mechanism responsible for generating time series of variable lengths.
	
	\subsection{Variations in frequencies of observation relative to the fundamental signal}
	\label{subsec:sampling rate}
	If we consider the time series as a sampling over time of some ideal or prototypical signal, then it is possible for that sampling of the ideal signal to vary in frequency from series to series.
	In the following, we describe how different sampling frequencies can lead to time series of unequal lengths.
	
	\subsubsection{Constant variations in the frequencies of the time series}
	\label{subsubsec:constant rate}
\begin{figure}
		\centering
		\begin{subfigure}[]{0.49\linewidth}
			\centering
			\includegraphics[width=\linewidth]{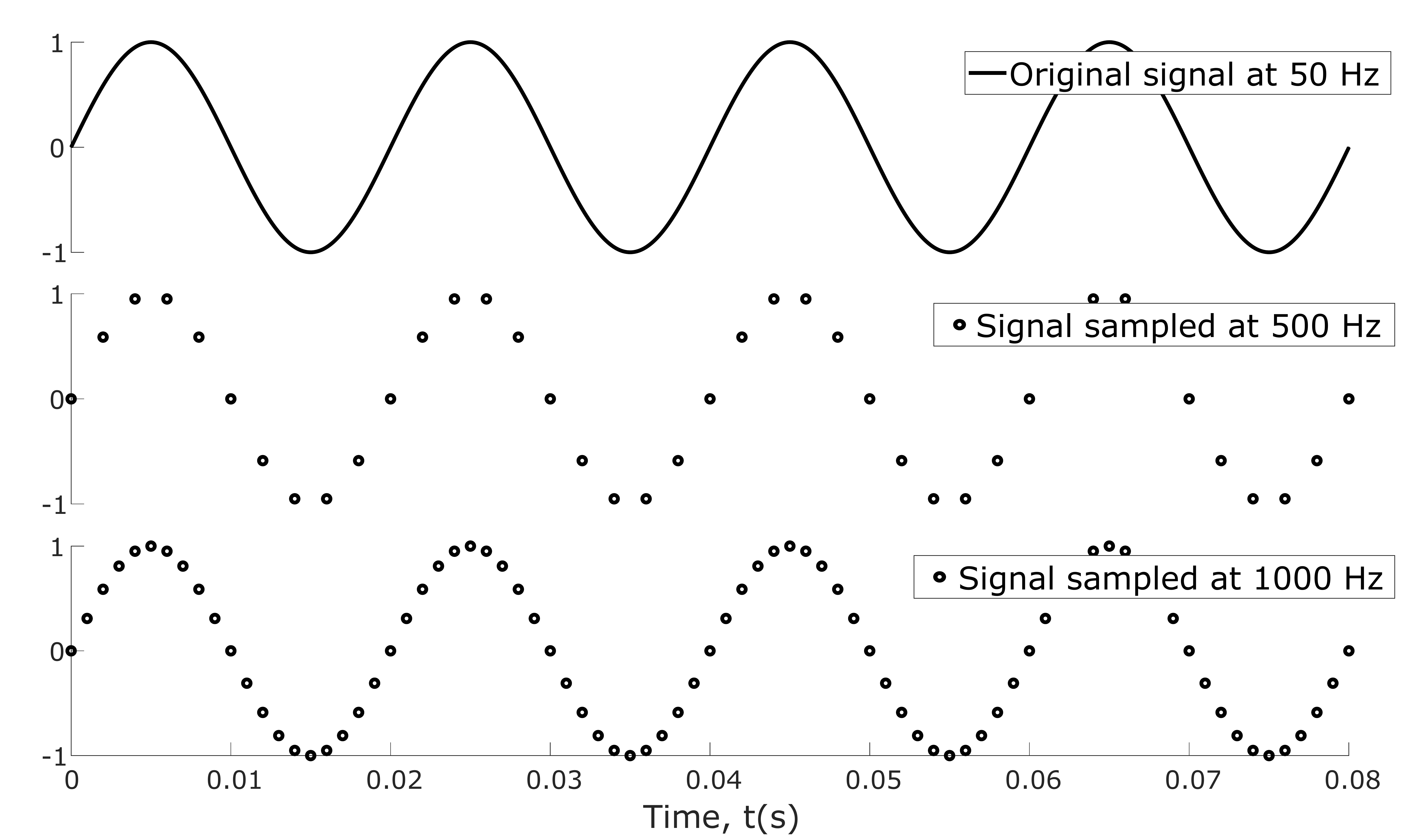}
			\caption{}
			\label{fig:sampled signal}
		\end{subfigure}
		\hfill
		\begin{subfigure}[]{0.49\linewidth}
			\centering
			\includegraphics[width=\linewidth]{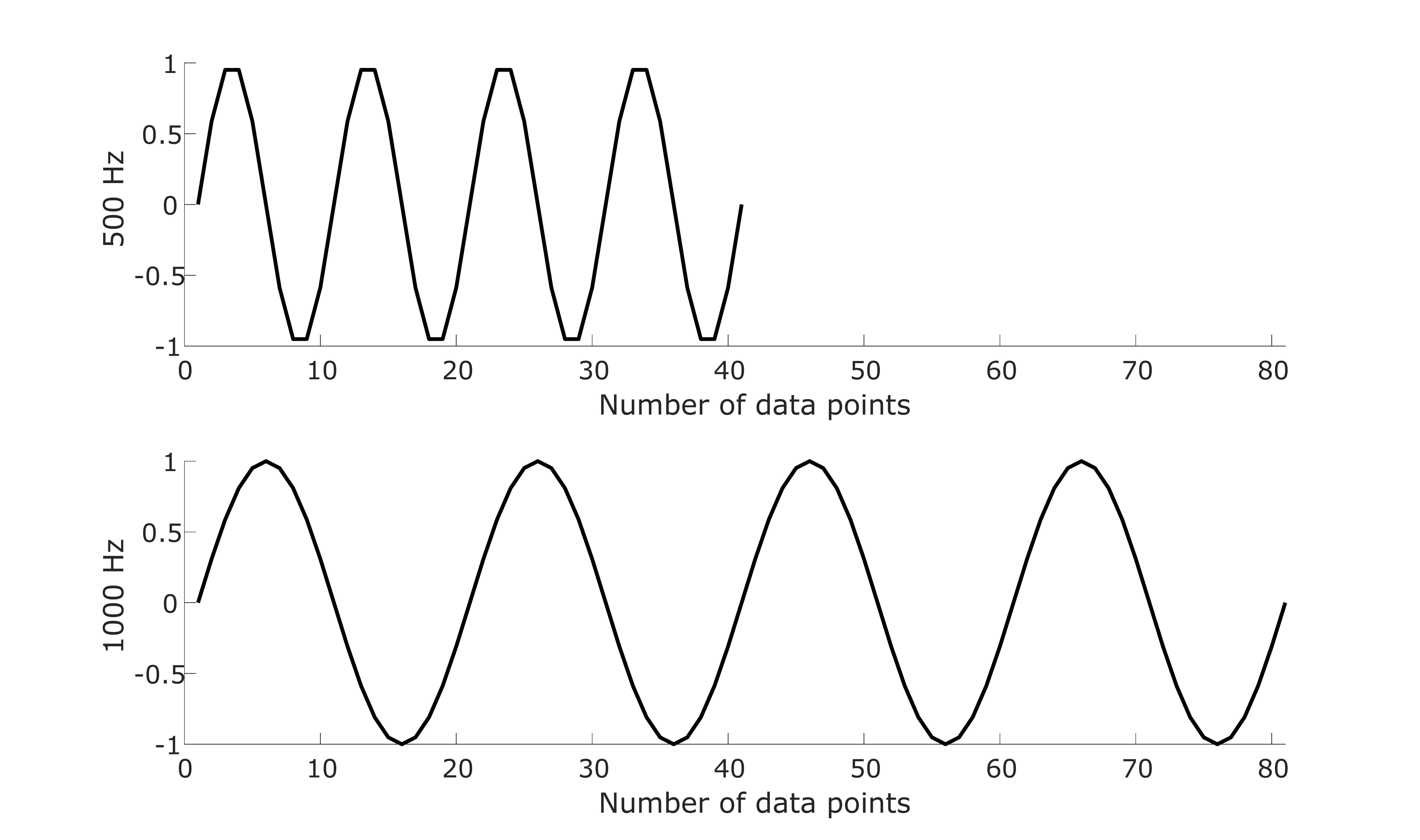}
			\caption{}
			\label{fig:sampled signal as time series}
		\end{subfigure}
		\caption{The top figure in (a) shows the original signal generated with 50 Hz frequency, followed by the series sampled with 500 Hz and 1 kHz respectively. The 1kHz series has twice as many data points as the 500 Hz series. Figure (b) shows both series in vector form, highlighting the unequal lengths of two time series drawn from the same signal.}
	\end{figure}

	Many time series data, such as voltage and temperature readings, are captured by sensors.
	Consider a toy example using a voltage signal generated from a 50 Hz frequency.
	The signal can be measured using a sensor with either 500 Hz or 1 kHz sampling frequency and we will still get the same 50 Hz sinusoidal voltage signal.
	This means that even though they are derived from the same signal, series A, sampled at 500 Hz, will have half as many points as series B, sampled at 1 kHz, illustrated in Figure \ref{fig:sampled signal}.
	Hence, as illustrated in Figure~\ref{fig:sampled signal as time series}, series A will be shorter than series B, while still measuring the same process.
	
	Another process by which time series data can be generated is by tracing the shape of an object in an image.
	\citep{ye2009time} and \citep{keogh2009supporting} show how two-dimensional images of leaves, skulls, projectile points, butterflies or shields can be converted into one dimensional time series data by tracing the edges of the object in the image.
	Video snippets can also be converted into time series data as shown by the famous Gun-Point problem \citep{ratanamahatana2004making,UCRArchive2018}, 
	where the goal is to identify if a person is holding a gun or pointing his/her fingers.
	
	These images and videos, when converted into time series, can have variations that affect their length.
	If two images of the same object are captured using cameras of different resolution, then the time series generated from the higher resolution camera will be longer. 
	This is because more pixels of the object will be used to form the time series. 
	The distance between the camera and the object will also affect the series length.
	If the camera is nearer to the object, 
	the image will have more pixels per object, which leads to a longer time series.
	If the camera is further away, a shorter time series is formed.
	
\begin{figure}
		\centering
		\begin{subfigure}[]{0.3\linewidth}
			\centering
			\includegraphics[width=\linewidth]{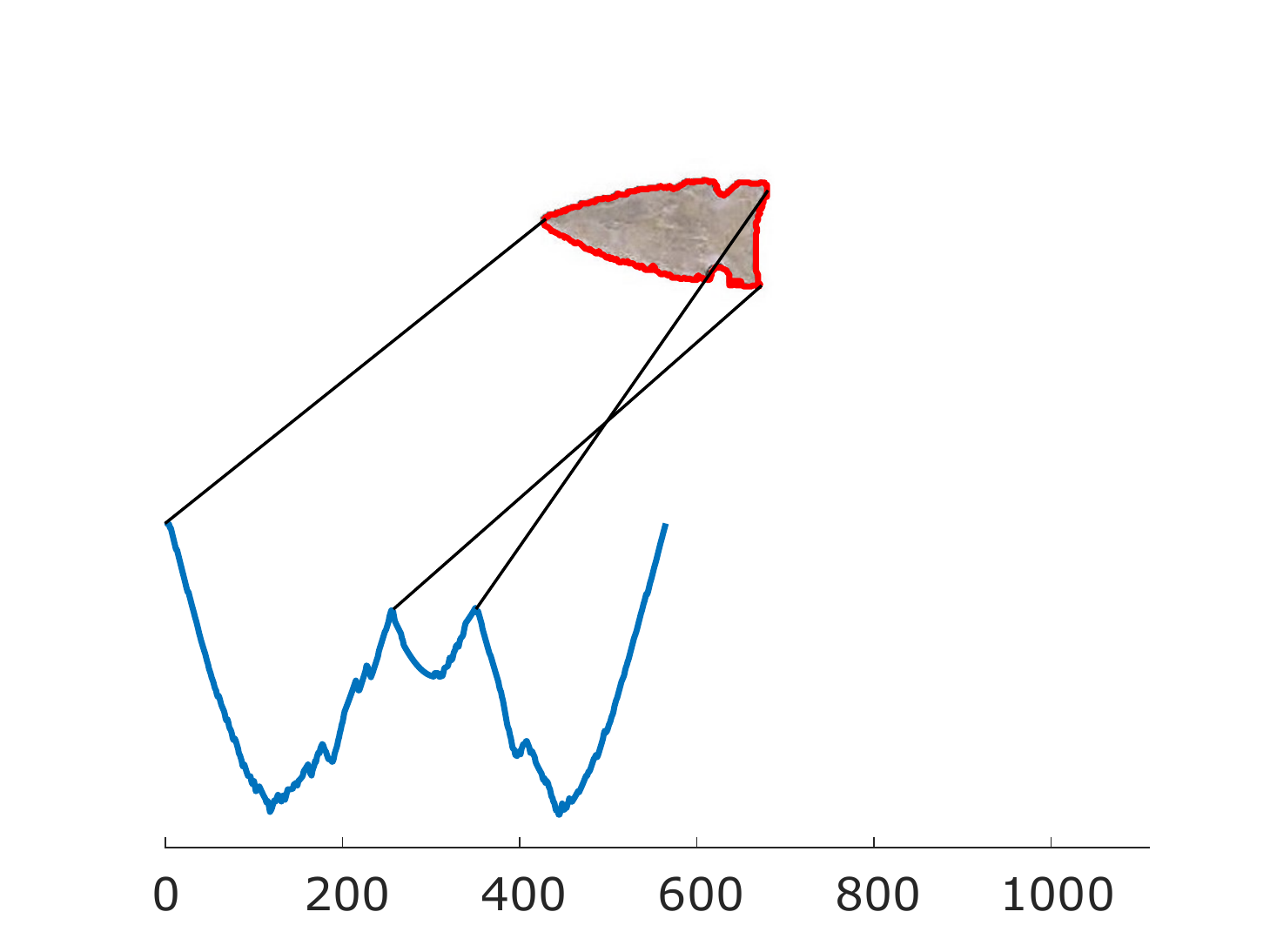}
			\caption{}
			\label{fig:arrowhead small}
		\end{subfigure}
		\hfill
		\begin{subfigure}[]{0.3\linewidth}
			\centering
			\includegraphics[width=\linewidth]{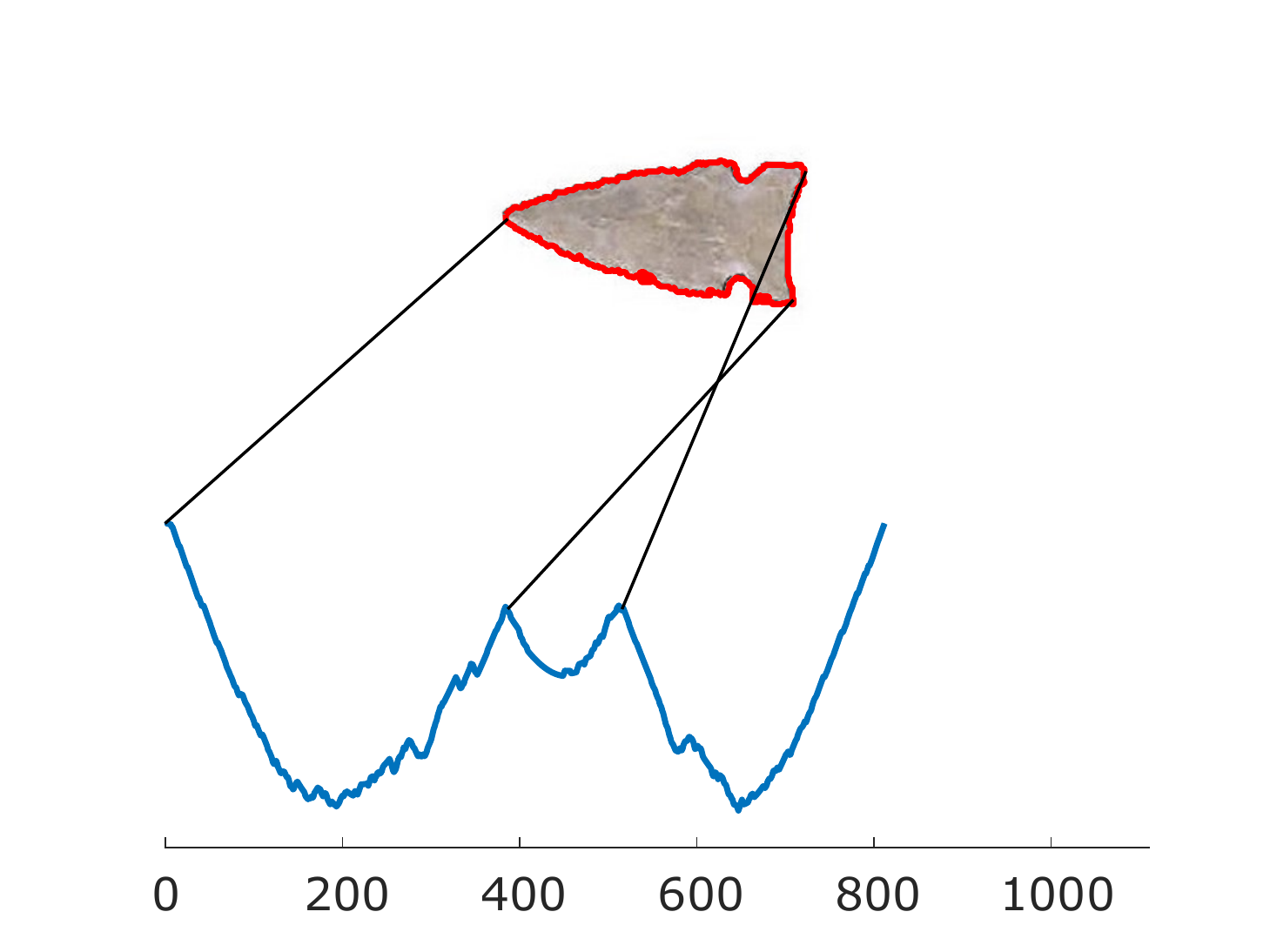}
			\caption{}
			\label{fig:arrowhead medium}
		\end{subfigure}
		\hfill
		\begin{subfigure}[]{0.3\linewidth}
			\centering
			\includegraphics[width=\linewidth]{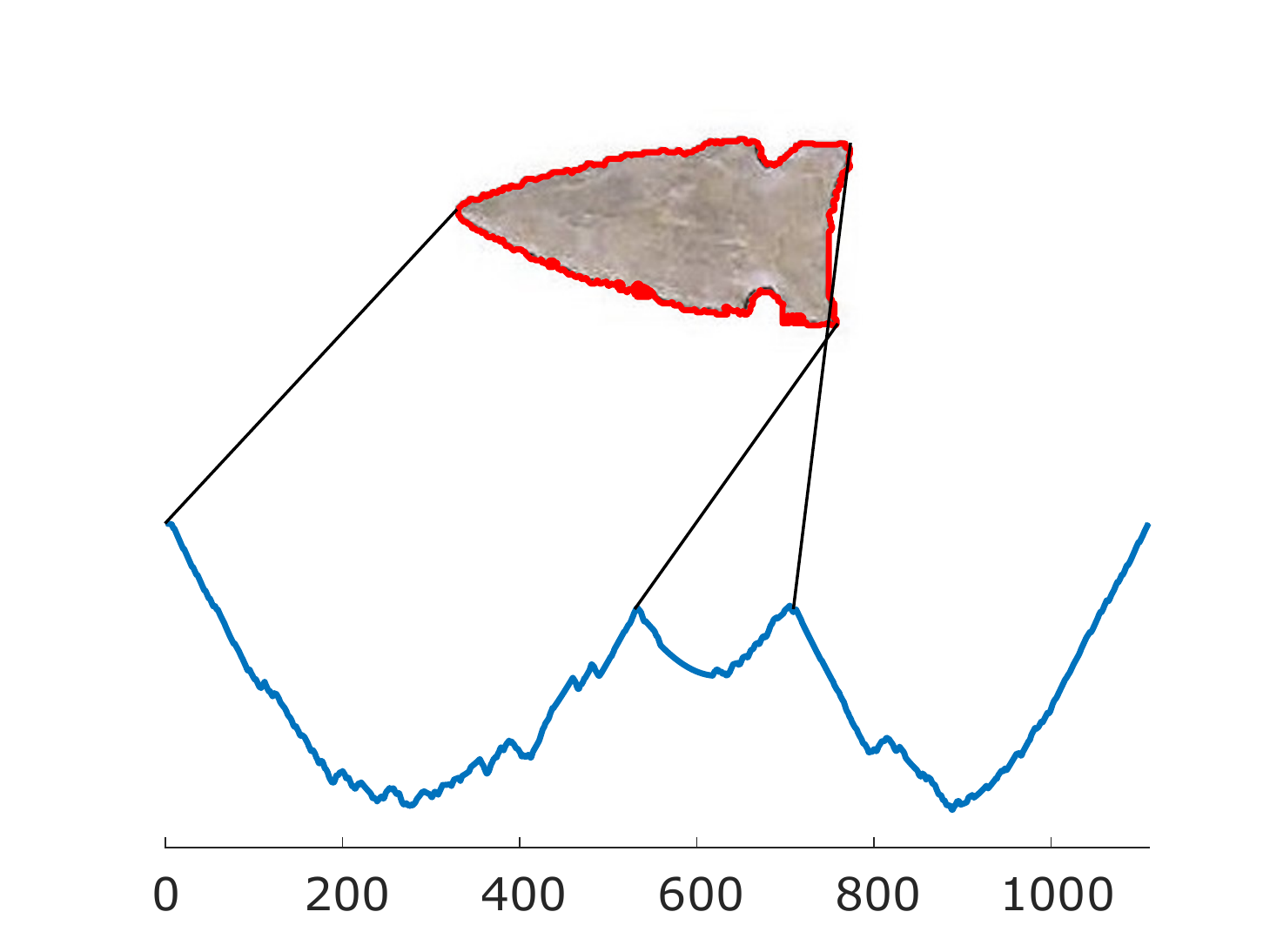}
			\caption{}
			\label{fig:arrowhead large}
		\end{subfigure}
		\hfill
		\begin{subfigure}[]{\linewidth}
			\centering
			\includegraphics[width=\linewidth]{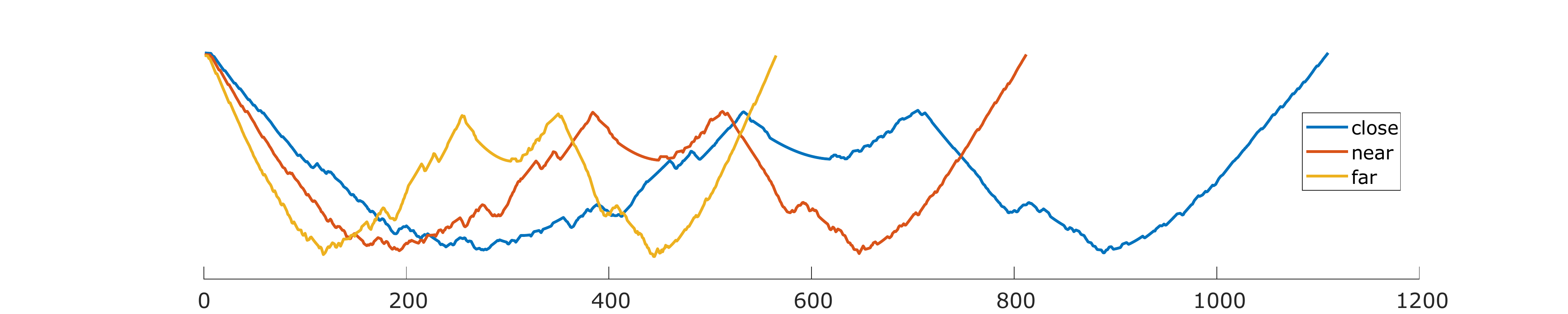}
			\caption{}
			\label{fig:arrowhead series}
		\end{subfigure}
		\caption{Image of the projectile point and time series formed when (a) the camera is far away, (b) near and (c) close to the object. The length of the time series in (a), (b) and (c) are 565, 812 and 1110 respectively, shown in (d). This clearly illustrates that the distance of the camera to the object can lead to different lengths time series.
		The projectile point image is taken from the supplementary website of \citep{ye2009time} and zoomed to simulate the effects of the different camera distances.}
		\label{fig:camera distance}
	\end{figure}

	Figure \ref{fig:camera distance} illustrates this scenario, where the image of a projectile point is taken at different distances to the camera, far (Figure \ref{fig:arrowhead small}), near (Figure \ref{fig:arrowhead medium}) and close (Figure \ref{fig:arrowhead large}).
	The image is then being converted into a one dimensional time series using the technique in \citep{keogh2009supporting}.
	The figures clearly show that time series formed at the different camera distance have different lengths.
	Since the camera's orientation is maintained from image to image, this scenario is equivalent to two time series having a constant factor in their frequencies relative to one another.

	\subsubsection{Variations in frequency}
	\label{subsubsec:varying rate}
	In the previous section, we have described time series of different lengths having a constant factor in their frequencies relative to one another.
	However in many real world applications, the relative frequency in a time series varies throughout the series.
	Recall that a time series are observations of process happening over a period of time.
	The duration of this process can be different relative to other time series.
	For example, the duration of a word spoken by two different people can be different.
	Person A may say ``appllle'' while person B may say ``aaaple'', where the ``l'' part in person A is longer than person B and the ``a'' part is longer in person B than person A.
	A familiar example of this is Southern American English (drawl) in which vowels tend to be pronounced longer than in northern American accents.
	
\begin{figure}
		\centering
		\includegraphics[width=0.5\linewidth]{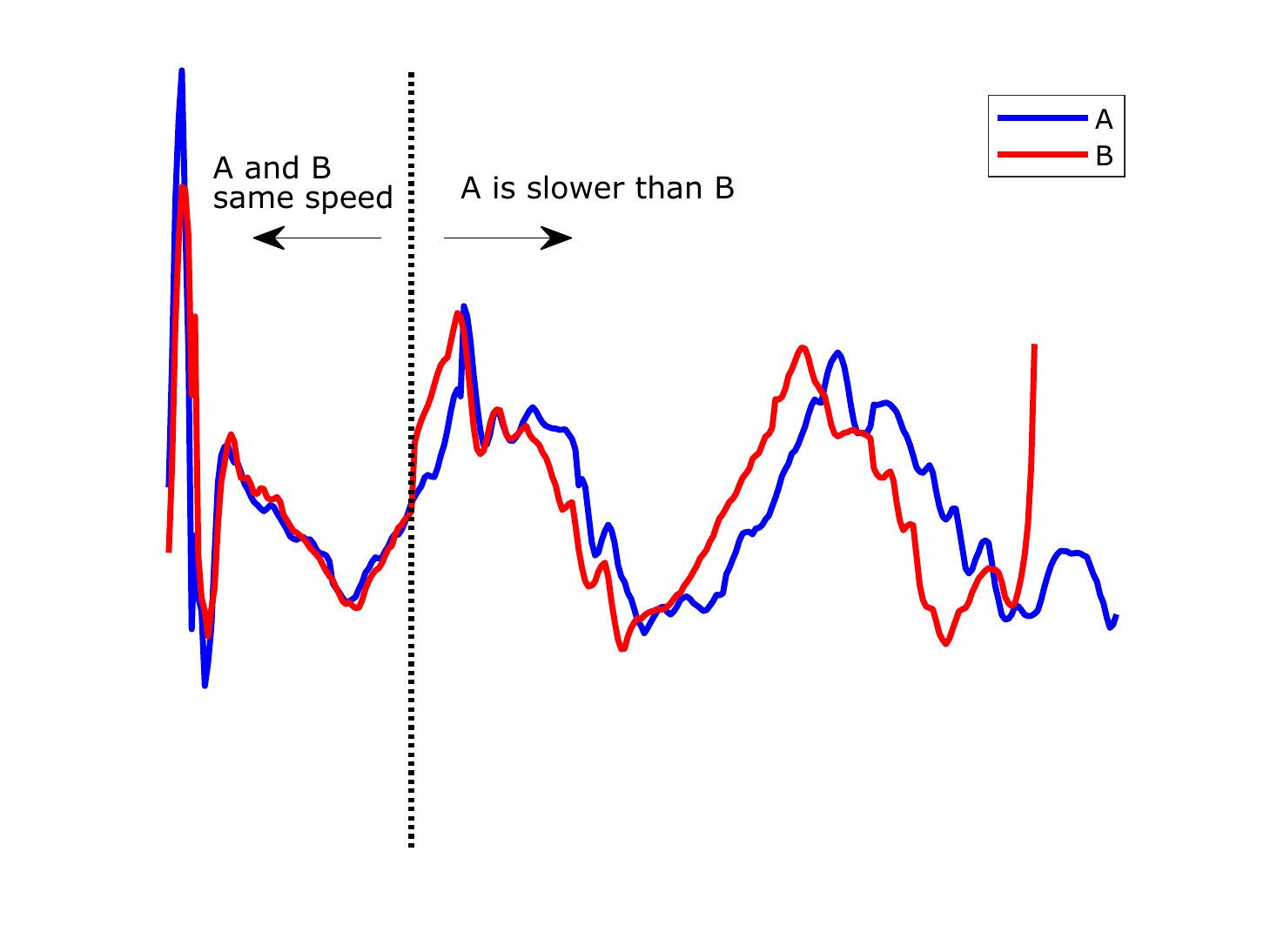}
		\caption{The `hu2' (class 6) gesture performed by two users A and B in the \texttt{GesturePebbleZ1} dataset from \citep{UCRArchive2018}}
		\label{fig:gesture}
	\end{figure}

	Another example of this form of variation is in gesture recognition applications \citep{mezari2017gesture}.
	One of the most recent additions to the UCR Archive \citep{UCRArchive2018}, the \texttt{GesturePebble} dataset, is from a study of the application of smart watches as a motion sensor for gesture recognition \citep{mezari2017gesture}.
	The data is collected using the 3-axis accelerometer Pebble smart watch.
	The labels are the six gestures performed by a person and the z-axis reading is then used to create the time series dataset \citep{UCRArchive2018}.
	It is very common for the gesture duration to vary, even if it is executed by the same person. 
	For instance, the person may unintentionally accelerate his/her hand at the end of the gesture, reducing the gesture duration.
	Figure~\ref{fig:gesture} illustrates this exact behaviour with two \texttt{GesturePebble} series
	for the same gesture. 
	The gesture for both users has the same speed at the start, then user B starts to accelerate relative to A, generating two time series of different lengths.
	
	\subsection{Variations to the start and/or end of the time series relative to another time series}
	\label{subsec:start and end}
	Time series may also differ in length due to missing observations. Missing observations within a series affect the relative sampling rate. However, missing observations  at the start and end give rise to a fundamentally different effect.
	A~time series with missing observations at the start gives us only the suffix; missing observations at the end gives us the prefix; and missing observations at both the start and end gives us a subsequence of the time series.
	Consider a toy example in speech recording.
	A person recording a speech may start or end the recording at different times; or the subject may unintentionally speak before the recording starts;
	all of which leads to incomplete recorded speeches with different starting and ending timings and series of various lengths.
	
\begin{figure}
	    \centering
	    \begin{subfigure}[]{0.3\linewidth}
			\centering
			\includegraphics[width=\linewidth]{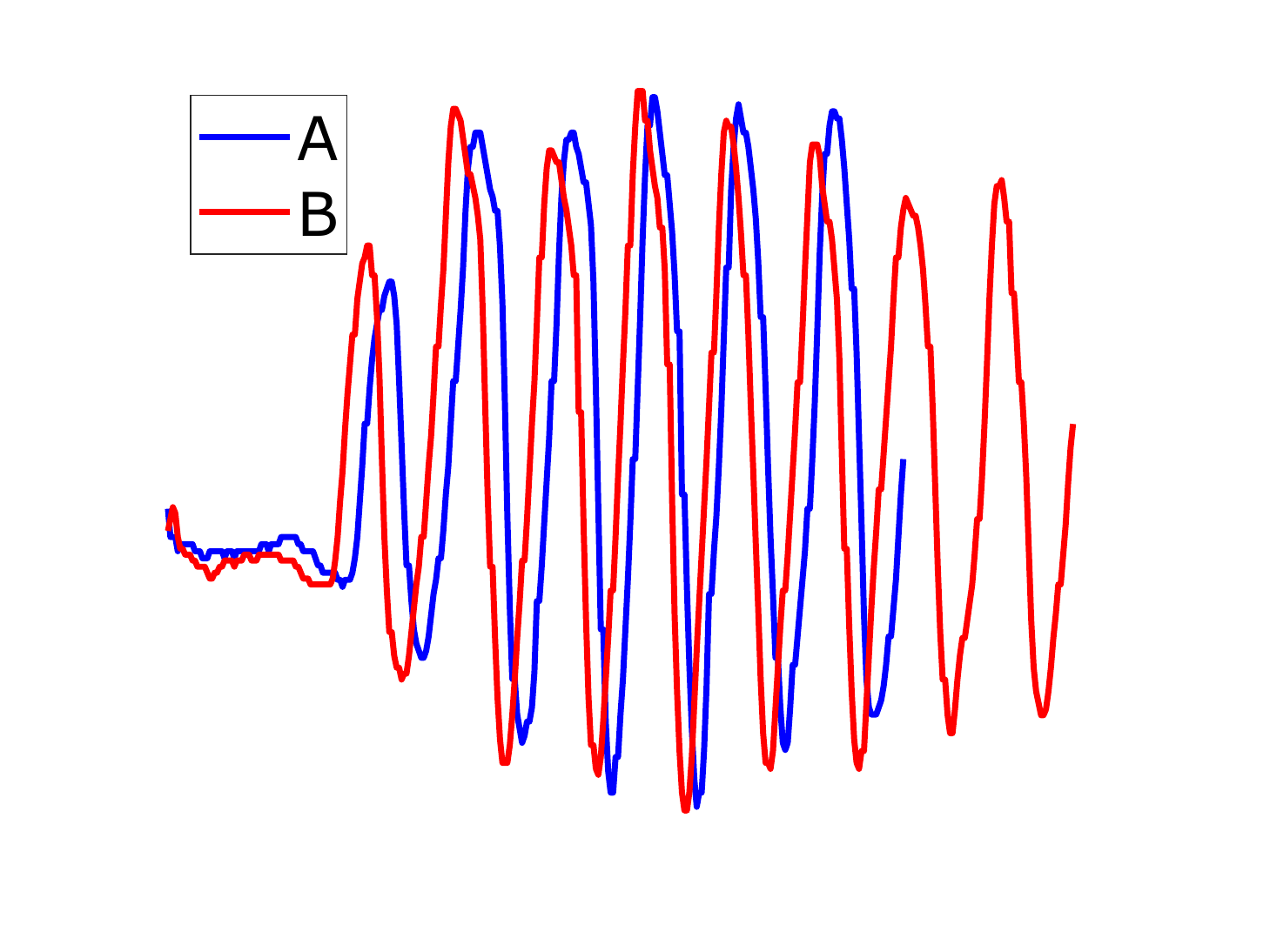}
			\caption{}
			\label{fig:prefix example}
		\end{subfigure}
		\hfill
	    \begin{subfigure}[]{0.3\linewidth}
			\centering
			\includegraphics[width=\linewidth]{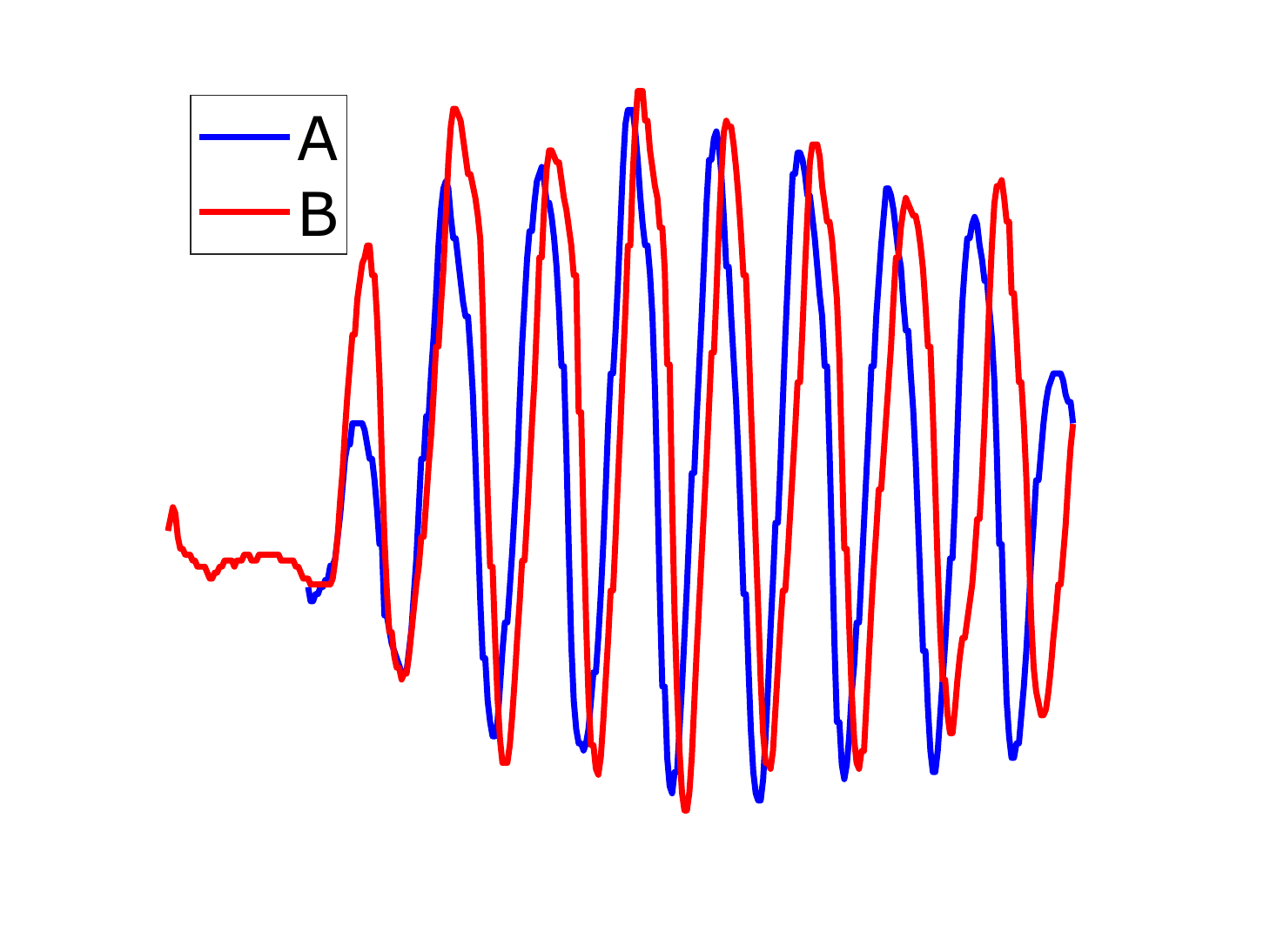}
			\caption{}
			\label{fig:suffix example}
		\end{subfigure}
		\hfill
		\begin{subfigure}[]{0.3\linewidth}
			\centering
			\includegraphics[width=\linewidth]{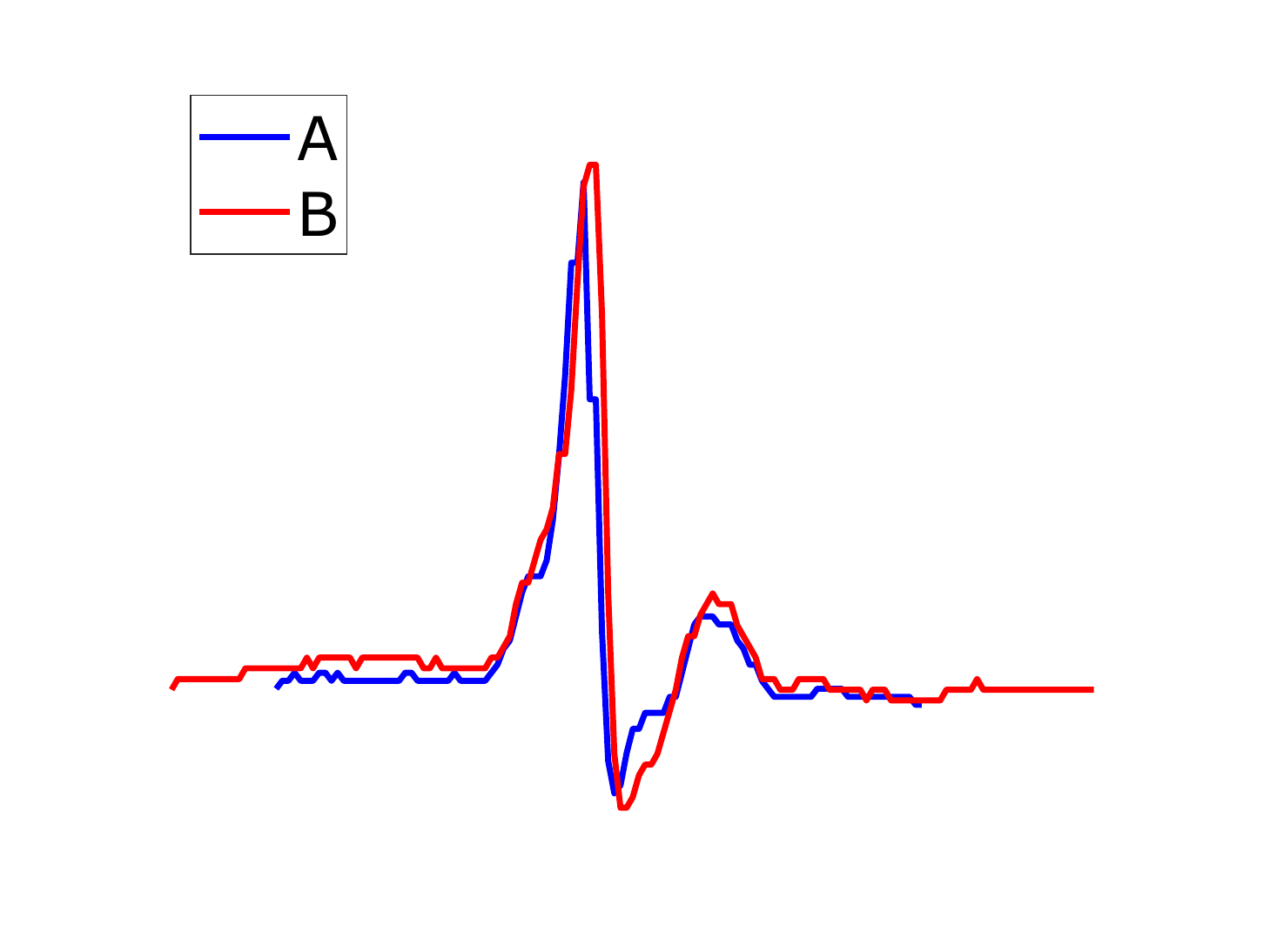}
			\caption{}
			\label{fig:subsequence example}
		\end{subfigure}
	    \caption{Example of (a) Prefix time series where there are missing observations at the end of series A, (b) Suffix time series where there are missing observation at the start of series A, (c) Subsequence time series where there are missing observations at the start and end of series A. These time series are taken from one of the variable lengths datasets, \texttt{AllGestureWiimoteY}, of the latest UCR time series archive \citep{UCRArchive2018}.}
	    \label{fig:prefix suffix subsequence examples}
	\end{figure}

	Figure~\ref{fig:prefix suffix subsequence examples} shows some examples of time series with variations at the start and/or end of the time series relative to another time series, taken from one of the variable length datasets of the latest update to the UCR time series archive, \texttt{AllGestureWiimoteY}  \citep{UCRArchive2018}.
	Figures~\ref{fig:prefix example} and \ref{fig:suffix example} show time series with missing observations at the end (so that the remaining time series is a prefix) and start (suffix) of series A relative to B, respectively. 
	Figure~\ref{fig:subsequence example} then shows the example of subsequence type time series with missing observations at the start and end of series A. 
	Note that these are time series obtained from real world sensors \citep{UCRArchive2018}.
	
	\section{Processing techniques for time series of unequal lengths}
	\label{sec:processors}
	In Section~\ref{sec:generators}, we introduced two classes of mechanism that generate time series of unequal lengths. 
	Many (if not most) TSC systems require all of the time series to be equal in length and are unable to handle time series of different lengths.
	In~consequence, it is common to preprocess time series to make them of equal length.
	In this section, we discuss some common preprocessing techniques and propose a new one.
	
	\subsection{Uniform scaling}
	\label{subsec:rescale}
	Uniform scaling is a common technique for equalizing time series lengths \citep{keogh2003efficiently}. It rescales one series to the length of the other.
	It is more common to stretch the shorter series than to shrink the longer, because shrinking entails data loss.
	This technique  has been  used in many time series tasks \citep{keogh2003efficiently,hu2013time,yankov2007detecting,gao2018efficient}. 
	Although widely used, it intuitively appears more applicable to series that differ in length due to varying frequencies relative to the underlying signal than to those that differ due to variations in start and end points, as illustrated in
	Figure~\ref{fig:prefix suffix subsequence examples}.
	
	\subsection{Low amplitude noise padding at the suffix of the time series}
	\label{subsec:suffix noise}
	Another common preprocessing approach is to add low amplitude noise to the suffix of the shorter time series \citep{UCRArchive2018}.
	This technique preserves the shape and does not change the information in the original series. 
	Although it has been used for benchmarking in the latest time series benchmark archive \citep{UCRArchive2018},
	it may not be suitable for some datasets and TSC algorithms, as illustrated in Figure~\ref{fig:dendrograms}. 
	Figure~\ref{fig:Comparison of suffix noise} in Appendix \ref{app:comparison suffix noise} plots the classification accuracy of \nndtw{} on the variable length datasets from the latest update to the UCR Archive \citep{UCRArchive2018} using the different processing techniques described in this section. 
	While it would be dangerous to generalize too far from this small collection of time series from similar domains, 
	these figures show that while padding the suffix with low amplitude noise works best overall, it is outperformed on at least two of the eleven datasets by each of the other approaches. 
	In particular, it does not work well for the \texttt{GestureMidAirD1} dataset, 
	where all the other techniques give better classification accuracy.
	
	\subsection{Low amplitude noise padding at the prefix and suffix of the time series}
	\label{subsec:prefix suffix noise}
	A direct extension to the previous technique is to add low amplitude noise at the prefix as well as the suffix of the shorter time series. 
	Similar to the previous technique, this technique preserves the shape and information of the time series. 
	This technique assumes that the longer time series has the complete pattern of an event that it is measuring and when the prefix and suffix of the shorter time series is padded with noise, the time series is ``shifted'' into the middle of the longer time series.
	Figure~\ref{fig:subsequence example} shows an example of such a shift with two series from the \texttt{AllGestureWiimoteY} dataset. 
	When both the prefix and the suffix of the shorter series, `user A,' are padded with random noise, the two series are well aligned. 
	
	\subsection{Single zero padding at the prefix and suffix of the time series}
	\label{subsec:prefix suffix zero}
	Similar to the previous technique, instead of padding the prefix and suffix with noise, we can pad it with a single zero. 
	However, unlike the previous techniques, it is important to be noted that this technique does not make the time series equal length.
	This is a new variant of prefix and suffix padding proposed in this work.
	It~is inspired by the fact that it is common practice to $z$-normalize each time series to $\mathcal{N}(0,1)$ before classification \citep{rakthanmanon2012searching,ratanamahatana2005three}.
	As the mean of the time series is zero, padding the prefix and suffix with a single zero allows \dtw{} to align any number of leading and trailing points to this padded point with minimum expected cost.
	It also reduces unintuitive alignments between the two time series by relaxing the \dtw{} constraint that the first and last point of the two time series must be aligned.
	The first and last point of the unequal length time series can be very different and if they are forced to be aligned,
	then the \dtw{} distance may be huge, which may lead to misleading results.

	
	\section{Direct processing of varying length time series}
	\label{sec:related work}
	Preprocessing to a standardized length, as discussed in Section \ref{sec:processors}, is necessary for TSC systems that cannot handle time series of various lengths.
	However, a number of TSC algorithms can handle time series of unequal lengths without preprocessing.
	
	Time series shapelets discovery \citep{ye2009time,mueen2011logical} is achieved by sliding the shapelet across the time series, 
	computing the Euclidean distance and returning the location in the time series that gives the lowest distance.
	This method has also worked well in many other applications such as time series motif discovery \citep{mueen2009exact} and time series data mining \citep{keogh2003efficiently, yeh2016matrix}.
	By definition, time series shapelets and motifs are shorter than a time series.
	All of these algorithms find the best match of a shorter subsequence with the longer time series. 
	This approach is called \emph{Subsequence Distance (SSD)}.
	It can be computed by sliding the shorter time series across the longer time series \citep{ye2009time}.
	At each step, the subsequence of the longer time series is renormalised and the Euclidean distance between the shorter time series and the subsequence of the longer time series is computed. 
	
	When used without a window, \dtw{} is directly applicable to time series of differing lengths.
	However, it is unclear how \dtw{} can be used to compare time series of unequal length when the warping path is constrained with a warping window.  
	If the series differ by more than the length of the warping window then the result is undefined. 
	Even if they do not, the suffix of the longer series is treated differently to the rest of the series, as each point has fewer choices of potential alignments.
	
	Uniform Scaling Distance is an alternative to \dtw{} that is also able to find the best match between two time series by stretching or shrinking one time series to match the other \citep{keogh2003efficiently}.
	The Euclidean distance is computed for all the possible ways in which the shorter time series can be stretched to the length of the longer time series.
	Then the best match is the one that gives the lowest Euclidean distance and is returned as the distance for uniform scaling. 
	Note that the scaled series from the Uniform Scaling distance can still have different lengths and it is different from the Uniform Scaling processing technique that scales all the time series in a dataset to the same length.
	\citep{keogh2003efficiently} demonstrated the importance of uniform scaling in many domains with unequal length time series, such as space shuttle telemetry monitoring, gene expression data and motion capture editing.
	Since then, Uniform Scaling Distance has been used in other time series mining tasks such as motif discovery \citep{yankov2007detecting, gao2018efficient}.
	In~this work, we use Uniform Scaling Distance with the classic \nn{} classifier and call it \nnus{}. 

    The Shape-based distance (\sbd{}) is a distance measure introduced in \citep{paparrizos2015k} for time series clustering.
    \sbd{} finds the best match between two time series using the cross-correlation measure. 
    The idea is similar to the \emph{Subsequence Distance} used in shapelets discovery \citep{ye2009time}.
    The cross-correlation measure keeps one time series static and slides another over the static one.
    Then the inner product of each shift is computed.
    The shift that gives the highest inner product is the best match and the highest inner product is returned as the distance of \sbd{}.
    We refer interested readers to the paper \citep{paparrizos2015k} for the technical details of \sbd{}.
    Here, we use the \sbd{} distance measure with the classic \nn{} classifier and name it \nnsbd{}.
    
	\section{Experimental evaluation}
	\label{sec:experiment}
	We performed two sets of experiments to understand the interactions between different mechanisms for generating time series of varying length, alternative preprocessing techniques and different strategies for TSC.
	First, we performed an experiment by modifying the 2015 UCR time series archive \citep{ucrarchive} to simulate the different mechanisms for generating time series of varying length, by generating a new version of each dataset for each generating mechanism. 
	Second, we evaluate all the techniques on the variable length datasets from the new UCR archive \citep{UCRArchive2018}.
	
	Due to time and resource constraints, 
	we do not evaluate all the state-of-the-art TSC algorithms.
	Instead, we select the following algorithms that could scale reasonable well as the representative of the major strategies for TSC. 
	We expect algorithms using similar strategies to have similar interactions with the mechanisms for generating varying lengths and preprocessing methods.

	\begin{enumerate}[label=\Alph*]
		\item \textbf{Classic similarity-based TSC algorithms} 
		\begin{enumerate}[label=\arabic*]
			\item One nearest neighbour with Euclidean distance (\nned{})
			\item One nearest neighbour with \dtw{} distance (\nndtw{})
		\end{enumerate}
		\item \textbf{Shift-invariance algorithms}
		\begin{enumerate}[label=\arabic*]
			\item One nearest neighbour with Subsequence Distance (\nnssd{})
			\item One nearest neighbour with Uniform Scaling Distance (\nnus{})
			\item One nearest neighbour with Shape-based Distance (\nnsbd{})
		\end{enumerate}
		\item \textbf{Computationally efficient state-of-the-art ensemble algorithms}
		\begin{enumerate}[label=\arabic*]
			\item Bag of SFA Symbols (\boss{})
			\item Proximity Forest (\pf{})
		\end{enumerate}
	\end{enumerate}
    
    The \nned{} and \nndtw{} represent the classic TSC algorithms, as they have been widely used before the introduction of the state-of-the-art ensemble algorithms \citep{wang2013experimental,ding2008querying}. 
    Since \nned{} is unable to handle unequal lengths time series, the longer time series is truncated to match the shorter time series.
    In this work, \nndtw{} is used without the warping window constraint, due to the issues relating to applying a warping window to series of different lengths.
    
    We choose \boss{} and \pf{} to represent state-of-the-art ensemble techniques due to their computational efficiency.
    We use the implementation from \cite{bagnall2017great} for \boss{} and \cite{lucas2019proximity} for \pf{}. 
    For simplicity, the default parameters are used for these two ensemble classifiers.
	
		
		
	
	We study all the preprocessing techniques outlined in Section \ref{sec:processors} and compare them to without processing. 
    Overall, the processing techniques used in this work include \textbf{No Processing}, \textbf{Uniform Scaling}, \textbf{Suffix Noise}, \textbf{Prefix Suffix Noise} and \textbf{Prefix Suffix Zero}.
	
	We evaluated all the TSC algorithms on all applicable combinations of a generating mechanism and a processing technique. 
	The \nn{} algorithm with Sliding Euclidean distance is only applied with No Processing and Prefix Suffix Zero, as making the series equal length is the same as \nned{}.
	The available \boss{} and \pf{} implementations are unable to handle time series of unequal length. In consequence, they are only applied with Uniform Scaling, Suffix Noise and Prefix Suffix Noise. 
	In total, we study $28$ combinations of a mechanism for varying series length, a series processing technique and a classifier.
	
	\subsection{Experiments on the modified UCR archive}
	\label{subsec:overall results}
	In this section, we study and evaluate the performance of alternative preprocessing algorithms and TSC strategies on varying length time series generated under each of the mechanisms described in Section \ref{sec:generators}.
	
	\subsubsection{Data preparation}
	\label{subsec:data}
	The 2015 UCR time series archive \citep{ucrarchive} is modified to simulate the different mechanisms.
	This allows us to understand the interactions between each mechanism, preprocessing technique and TSC strategy.
	For each mechanism, for each of the train and test set of each dataset in the archive, 85\% of the instances are randomly selected and modified, leaving 15\% unchanged.
	In this work, we perform five modifications (\textbf{Uniform Sampling}, \textbf{Non-Uniform Sampling}, \textbf{Prefix}, \textbf{Suffix} and \textbf{Subsequence}) to the datasets. 
	The details of the modifications are described in Appendix \ref{app:modified ucr}.
	To ensure reproducibility of our experiments, our modified data and source code have been uploaded to \url{https://bit.ly/VaryLengthTSC}.
	
	
	\subsubsection{Evaluation}
\begin{figure}
	    \centering
	    \includegraphics[width=\textwidth,height=\textheight,keepaspectratio,trim={1.5cm 0 1.5cm 0},clip]{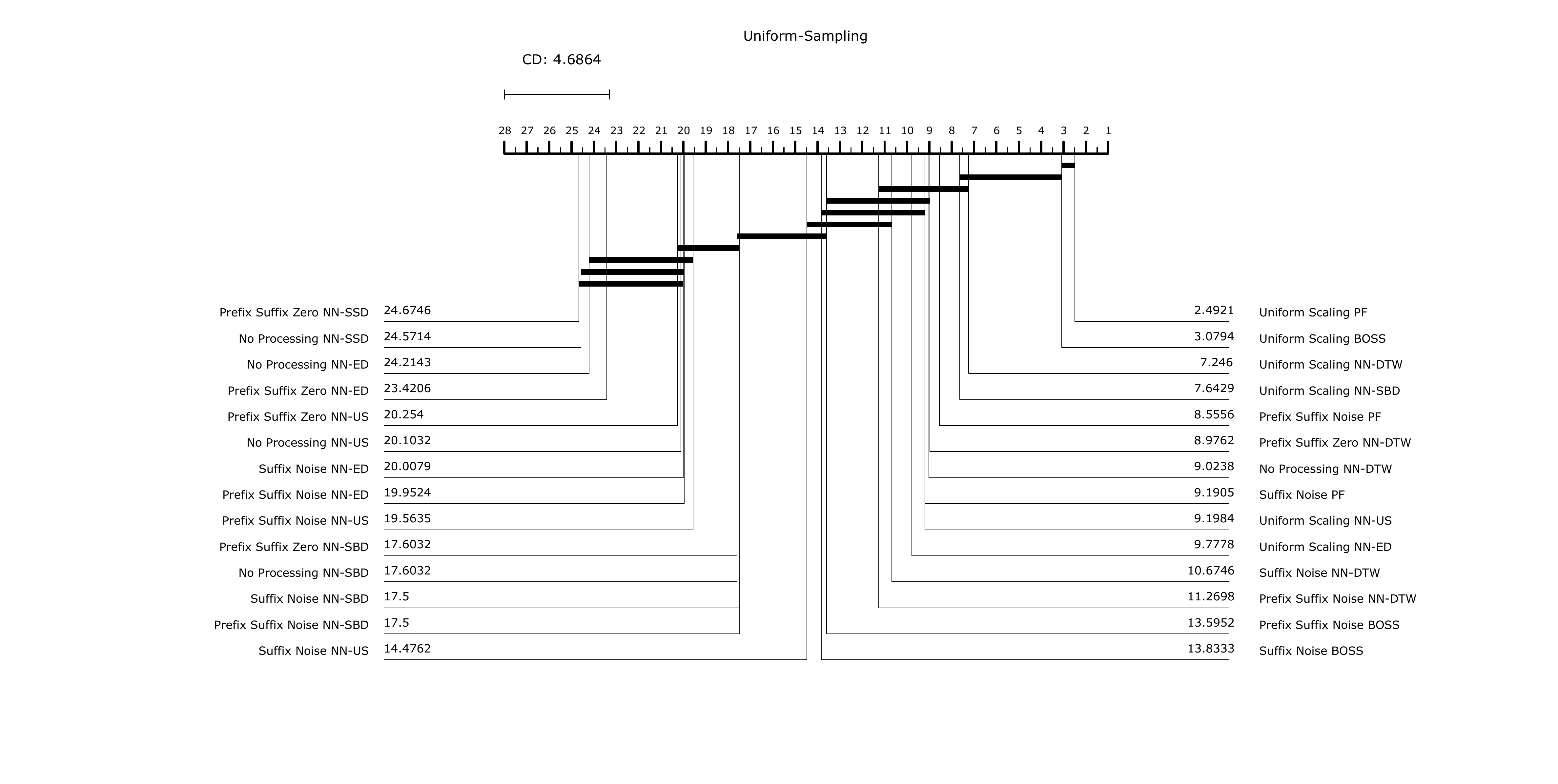}
	    \vspace{-40pt}
	    \caption{Rankings of each classifier processing technique pair on datasets generated by Uniform Sampling.}
	    \label{fig:CD uniform sampling}
	\end{figure}

\begin{figure}
	    \centering
	    \includegraphics[width=\textwidth,height=\textheight,keepaspectratio,trim={1.5cm 0 1.5cm 0},clip]{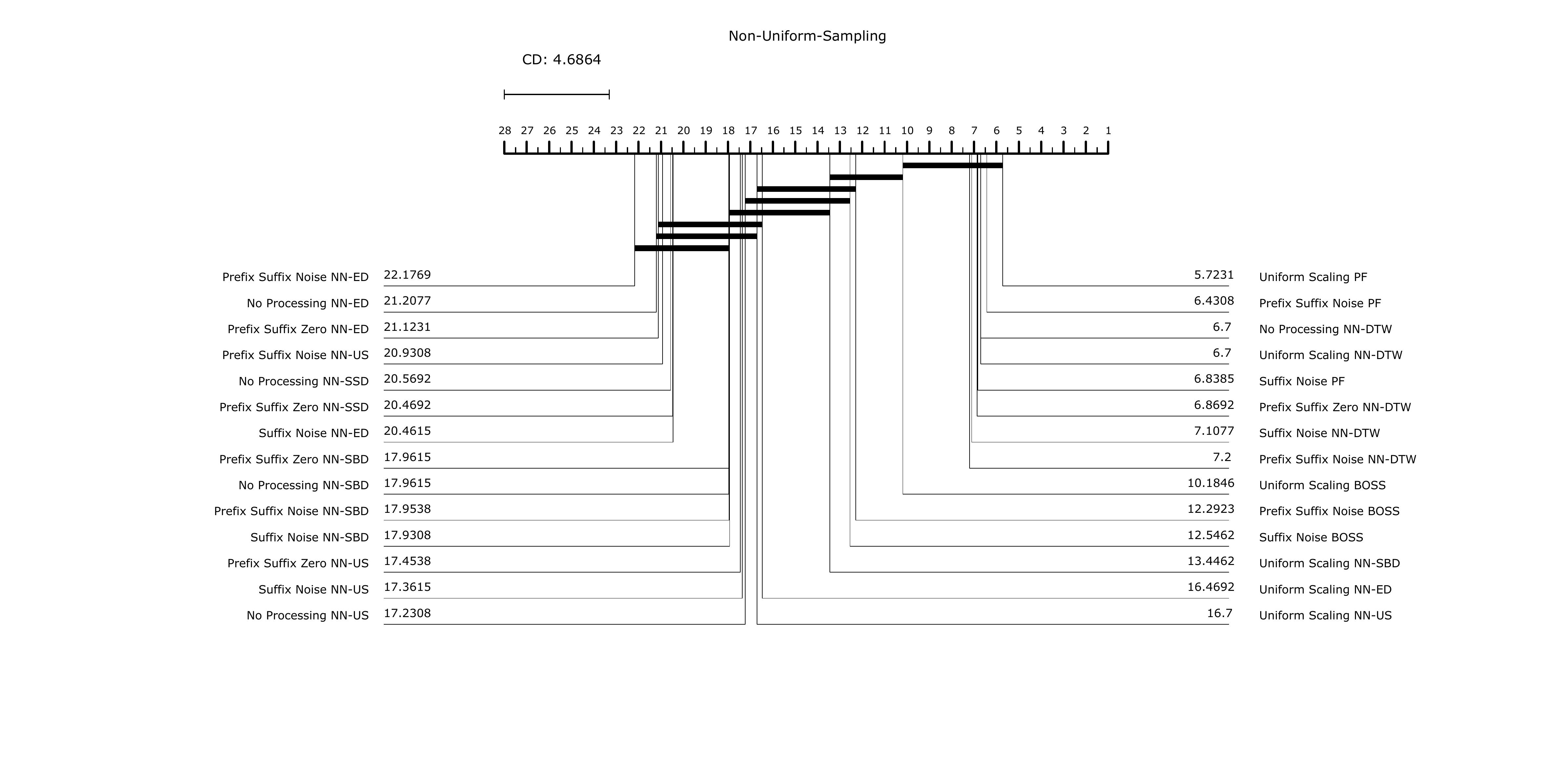}
	    \vspace{-45pt}
	    \caption{Rankings of each classifier processing technique pair on datasets generated by Non Uniform Sampling mechanism.}
	    \label{fig:CD non uniform sampling}
	\end{figure}

\begin{figure}
	    \centering
	    \includegraphics[width=\textwidth,height=\textheight,keepaspectratio]{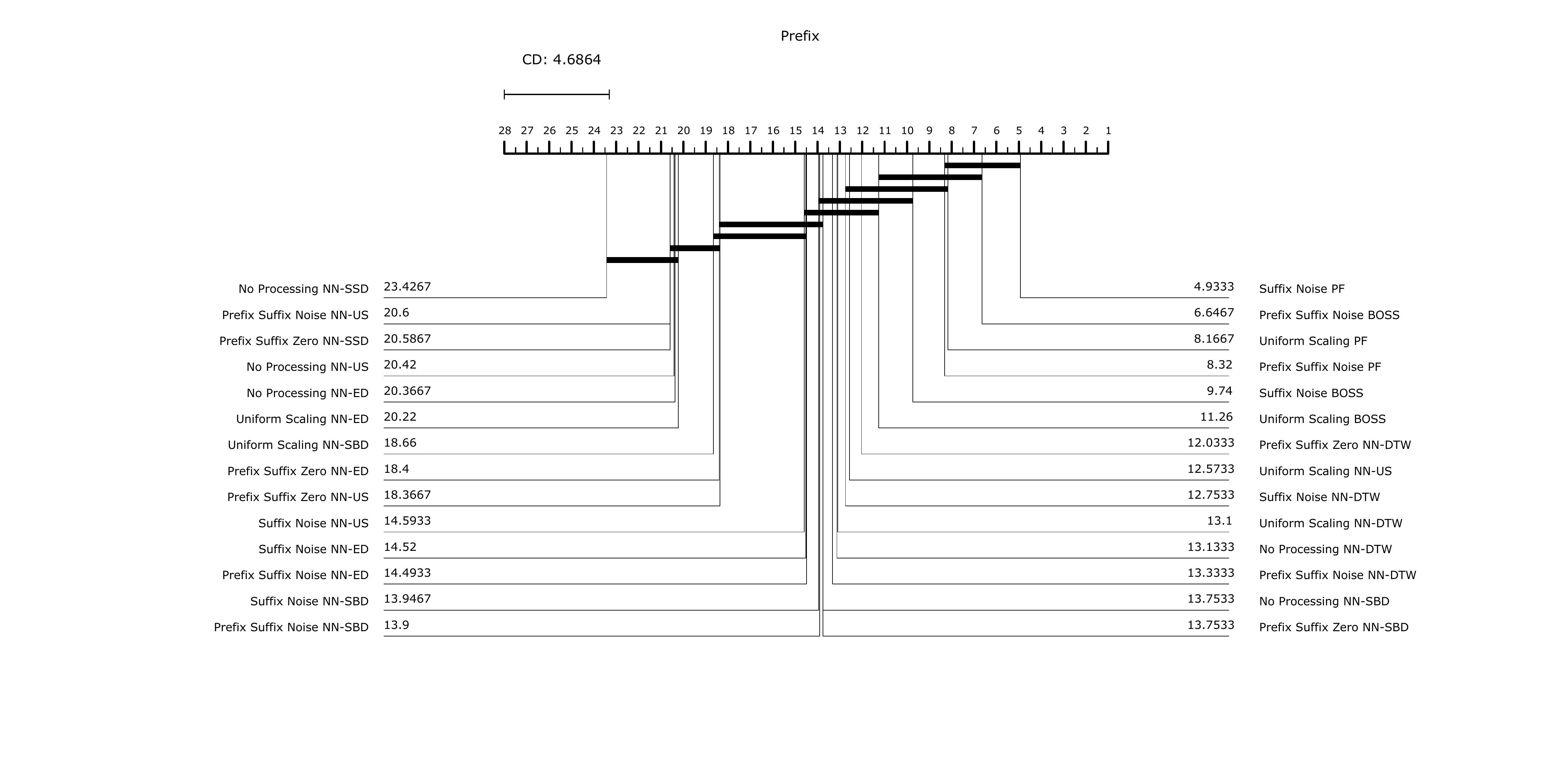}
	    \vspace{-40pt}
	    \caption{Rankings of each classifier processing technique pair on datasets generated by Prefix mechanism.}
	    \label{fig:CD prefix}
	\end{figure}

\begin{figure}
	    \centering
	    \includegraphics[width=\textwidth,height=\textheight,keepaspectratio,trim={1.5cm 0 1.5cm 0},clip]{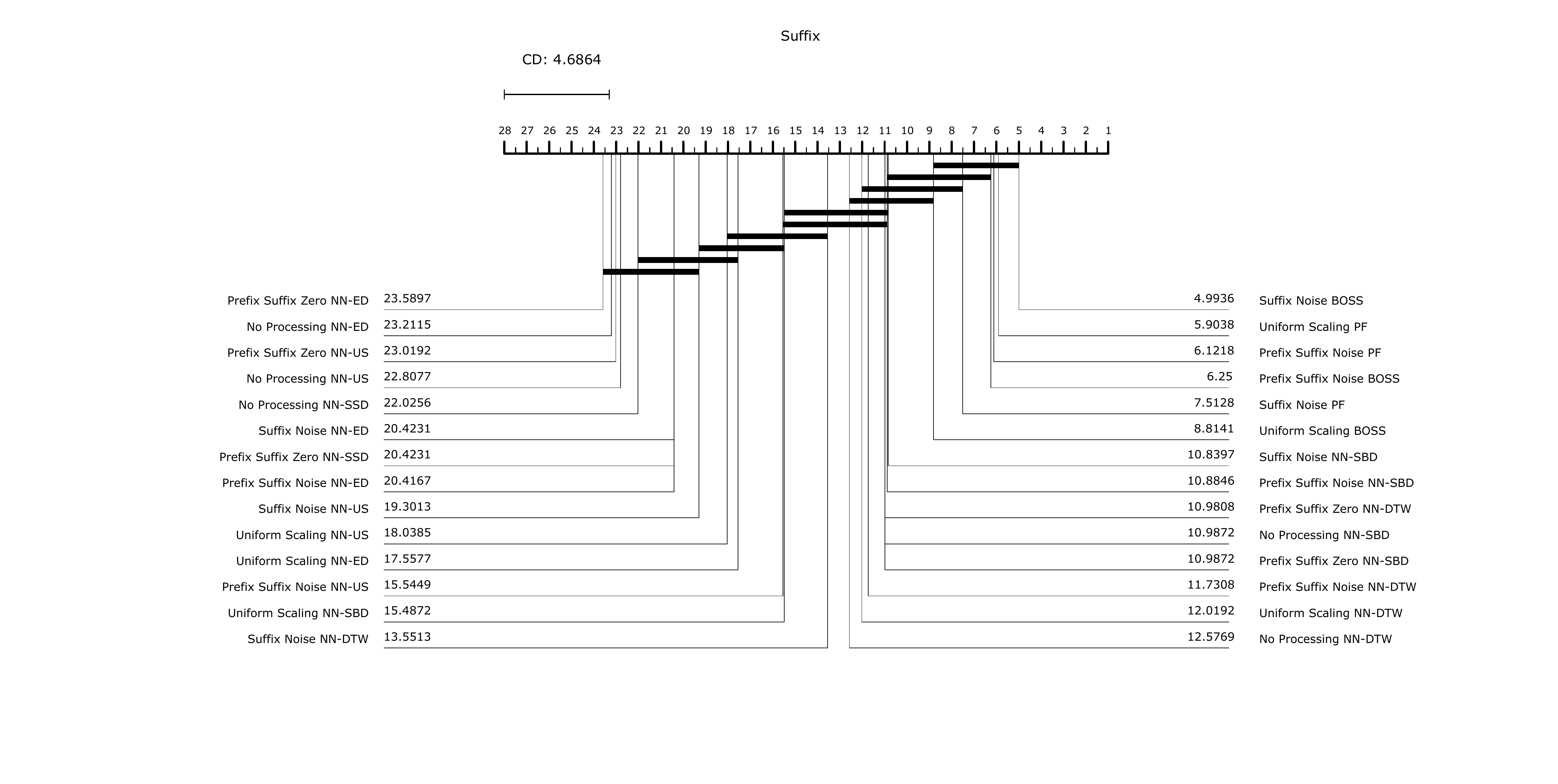}
	    \vspace{-40pt}
	    \caption{Rankings of each classifier processing technique pair on datasets generated by Suffix mechanism.}
	    \label{fig:CD suffix}
	\end{figure}

\begin{figure}
	    \centering
	    \includegraphics[width=\textwidth,height=\textheight,keepaspectratio,trim={1.5cm 0 1.5cm 0},clip]{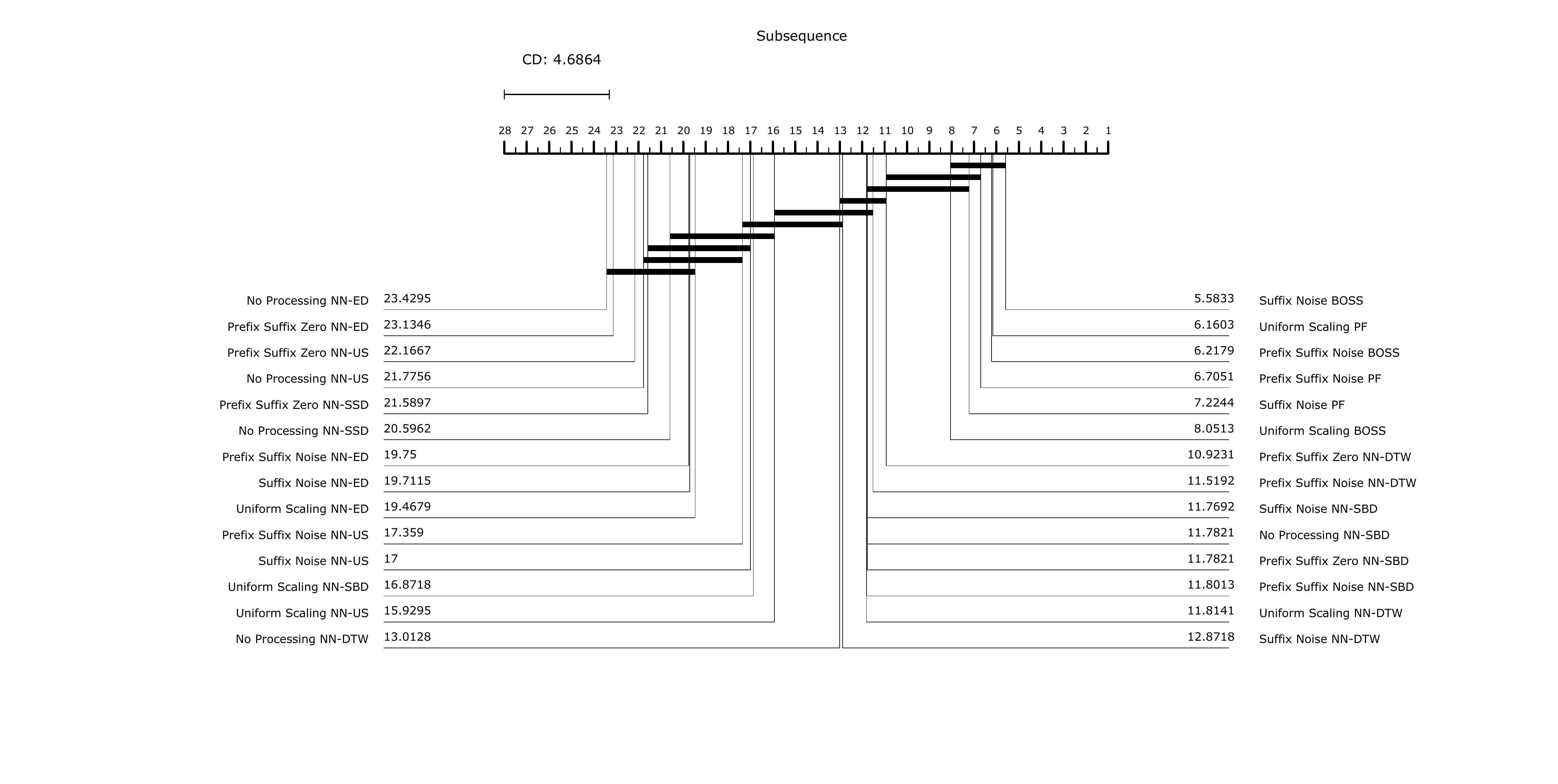}
	    \vspace{-45pt}
	    \caption{Rankings of each classifier processing technique pair on datasets generated by Subsequence mechanism.}
	    \label{fig:CD subsequence}
	\end{figure}

	We compute the classification accuracy for each  preprocessing-technique and TSC-classifier pair across all 85 UCR time series datasets. 
	For each dataset, we rank the preprocessor-classifier pair and compute the average rank across all datasets.
	Following the methodology in \citep{demvsar2006statistical}, when comparing $k=28$ alternatives over $N=85$ datasets, 
	at a level of $\alpha=0.05$, 
	the critical value $q_{0.05}=3.7145$ and 
	the average rank between two alternatives must be larger than the critical difference ($CD=4.6864$, calculated as follows) to be statistically significant.
	
	\begin{equation}
	    CD=q_{0.05}\sqrt{\frac{k(k+1)}{6N}}=3.7145\sqrt{\frac{28\cdot (29)}{6\cdot 85}}=4.6864
	\end{equation}
	The results are presented in the critical diagrams in Figures \ref{fig:CD uniform sampling}, \ref{fig:CD non uniform sampling}, \ref{fig:CD prefix}, \ref{fig:CD suffix} and \ref{fig:CD subsequence}.
	
	Figure \ref{fig:CD uniform sampling}
	shows the average rank for each preprocessor-classifier pair on time series data sampled with fixed frequencies. 
	Lower ranks indicate lower error, with the lowest average rank appearing rightmost in the figure.
	Pairs that are not significantly different from one another are grouped together by a black bar. 
	The top-ranked pair is Uniform Scaling with Proximity Forest.
	Overall, Uniform Scaling achieves a better rank than any alternative preprocessor when coupled with any classifier.
	No Processing performs almost as well as Uniform Scaling for \nndtw{}, which confirms the findings in \citep{ratanamahatana2005three}. 
	However, it is the worst or second worst alternative for all other classifiers that can handle varying length series, suggesting that for most classifiers, almost any preprocessing strategy is better than none for time series that differ due to differing fixed sampling frequencies.

	Figure \ref{fig:CD non uniform sampling} shows the average rank for each preprocessor-classifier pair on time series data generated with varying frequencies.
	The results suggest that for this mechanism, algorithms that use elastic distances, such as the Proximity Forest and \nndtw{}, outperform the other algorithms, including an ensemble that does not use elastic distances, such as \boss{}.
	This is because they are able to capture the varying frequencies in the time series by shrinking or stretching the time series.
	The results also indicate that there are no significant differences between the different processing techniques for Proximity Forest and \nndtw{}.
	Similarly to the case for uniform sampling, Uniform Scaling  performs well overall on any classifiers for time series generated from this mechanism, while other processing strategies are significantly worse for most classifiers.
	
	The average ranks for each of the combinations of processing technique and classifier on prefix, suffix and subsequence type time series are shown in Figures~\ref{fig:CD prefix}, \ref{fig:CD suffix} and \ref{fig:CD subsequence} respectively.
	Proximity Forest and \boss{} perform equally well on these three types of mechanisms without being significantly different, especially when the suffixes of the time series are padded with noise.
	The different processing strategies are not significantly different for \nndtw{}, demonstrating the robustness of \nndtw{}.
	Figure \ref{fig:CD prefix} indicates that padding the suffix with noise is either the best or second best technique on all classifiers for prefix type time series.
	Despite the intuitiveness of padding prefix type time series with noise, uniformly scaling them to the same length is also very competitive. 
	As shown in both Figures~\ref{fig:CD suffix} and \ref{fig:CD subsequence}, adding noise to the prefix and suffix of the time series helps improve the classification accuracy. 
	
	Furthermore, the results from these figures also show that our proposed technique, padding prefix and suffix with a single zero, works very well with \nndtw{}, regardless of the mechanism that generates the variations in length.
	However, as expected, it did not work well for other classifiers, as it was specifically designed for \dtw{}.
	
	The results strongly suggest that the choice of processing technique affects classification accuracy differently for series that vary in length for different reasons.
	For the uniform sampling mechanism in Figure \ref{fig:CD uniform sampling},
	Proximity Forest, \boss{} and \nnsbd{} are more accurate when the time series are uniformly scaled to the same length compared to the other processing techniques.
	It is also important to note that when the time series are uniformly scaled, \nndtw{} performs better than Proximity Forest with other processing techniques.
	For non uniform sampling mechanism, there are no significant differences between the processing techniques for each of the classifier, but it is clear that uniformly scaled to the same length works best. 
	Padding the suffix of a prefix time series with random noise is significantly more accurate for Proximity Forest, while padding both prefix and suffix is significantly more accurate for \boss{}. Unsurprisingly, uniformly scaled to the same length is significantly more accurate for \nnus{}.
    On the other hand, padding suffix with random noise is significantly more accurate for \boss{} on the suffix and subsequence mechanism, while it is the worst for Proximity Forest.
	These results are observed in the critical difference diagrams.
	Making time series equal in length is crucial for Euclidean distance based algorithms to achieve good performance, while it is not necessary for \nndtw{} and \nnsbd{}. 
	
	\subsection{Experiments on variable lengths datasets from the new UCR archive}
	So far, we have only evaluated the performance of each preprocessor-classifier pair on synthetic examples of individual mechanisms that may cause time series to vary in length. 
	However, real-world time series data may be affected by a combination of different mechanisms. 
	Hence, we performed another set of experiments on the eleven of the variable lengths datasets from the new UCR time series benchmark archive \citep{UCRArchive2018}. 
	These datasets are collected from real-world sensors for gesture recognition and have not being preprocessed. 
	Therefore, we follow the recommendations from \citep{UCRArchive2018} to process the data.
	First, we linearly interpolate the missing values in the data.
	Then, we process the data using each of the techniques described in Section \ref{sec:experiment}.
	Finally, the data are normalized to $\mathcal{N}(0,1)$.
	
\begin{figure}
	    \centering
	    \includegraphics[width=\textwidth,height=\textheight,keepaspectratio,trim={1.5cm 0 1.5cm 0},clip]{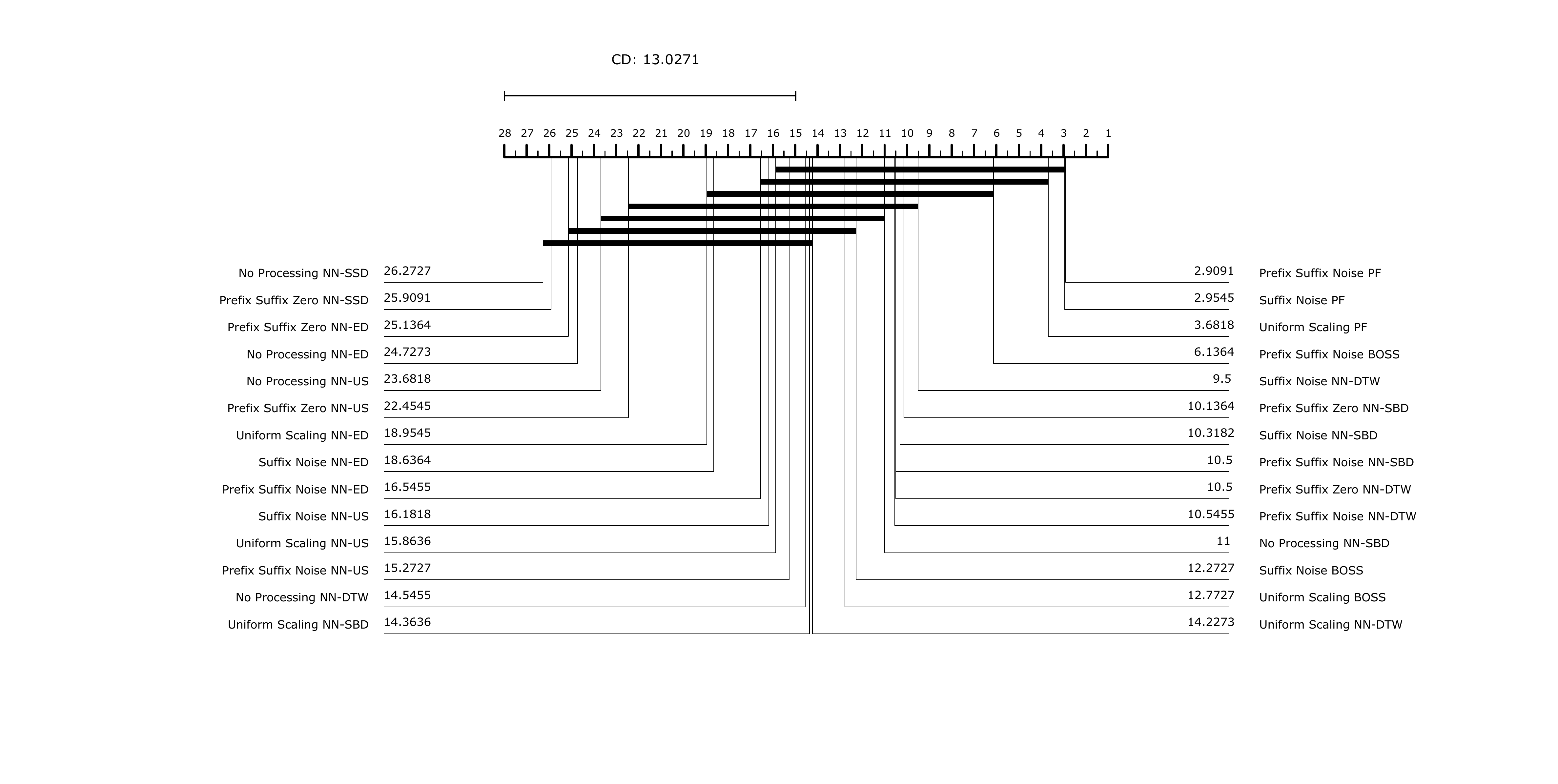}
	    \vspace{-40pt}
	    \caption{Rankings of each classifier processing technique pair on the variable lengths dataset from the latest UCR archive \citep{UCRArchive2018}.}
	    \label{fig:CD new UCR}
	\end{figure}

	Similar to the previous experiments, each preprocessor-classifier pair is ranked on classification accuracy for each dataset and the average rank is computed across all eleven datasets.
	Hence, with $k=28$ preprocessor-classifier pairs, $N=11$ datasets and a significance level of $\alpha=0.05$, the average rank has to be larger than a critical difference of $CD=13.0271$ to be significantly different. 
	Figure \ref{fig:CD new UCR} shows the average ranks for each preprocessor-classifier pair across the variable lengths datasets. 
	The results show that Proximity Forest is robust in dealing with variable length time series such as these, as it is the best performing classification algorithm regardless of the compatible preprocessing technique used. 
	\boss{} when paired with Prefix Suffix Noise is very competitive, but its performance drops when paired with other processing techniques. 
	As \boss{} performs better with Uniform Scaling when the series differ due to differing fixed sampling frequencies, this result is suggestive that none of these time series represent such data.
	The best processing technique for \nndtw{} is Suffix Noise, which is the technique used to benchmark the latest UCR archive \citep{UCRArchive2018}, followed by our proposed technique, Prefix Suffix Zero.
	An interesting result is that \nnsbd{} is also very competitive, which was not observed in the previous experiment with synthetic data. 
	Finally none of the Euclidean distance based algorithms performed well, with Sliding Euclidean Distance being the worse.
	
	\section{Conclusion and future work}
	\label{sec:conclusion}
	In this work, we have studied how the mechanisms that cause time series to differ in length affect the classification accuracy of TSC algorithms. 
	We have proposed two fundamental mechanisms for generating time series of differing lengths --- varying sampling frequencies and varying start and end points. 
	We identify two types of varying frequency --- fixed frequency and varying frequency. 
	We identify three types of differing start and end point ---  prefixes, suffixes and subsequences.
	
	We have shown empirically that different TSC algorithms and preprocessing techniques interact differently with different mechanisms for generating variations in length.
	More importantly, matching the right processing technique for an underlying mechanism is critical, but this matching varies for different types of TSC algorithm. 
	Our experiments conclude that distance-based ensembles, as represented in these experiments by Proximity Forest, coupled with Uniform Scaling provide competitive performance over a wide range of mechanisms for generating variability in series length. 
	\boss{} coupled with Prefix Suffix Noise is even more effective on some datasets, but appears less effective when the series are created with differing sampling rates.
	Nonetheless, the classic \nndtw{} is also a viable option for limited computational resources scenarios. 
	
	There are numerous directions for future work. It would be valuable to extend the study to further TSC algorithms. It remains an open problem how the most suitable processing technique might be selected for a given time series. We have identified two fundamental classes of mechanism that may cause time series to differ in length. It remains an open question whether there are further such processes and, if so, how they interact with different preprocessors and classifiers.
	
	\footnotesize
	\bibliographystyle{spbasic}
	\bibliography{reference}
	
	\clearpage
	\appendix
	\section{Comparison of classification accuracy using different processing techniques}
	\label{app:comparison suffix noise}
	
\begin{figure}[h]
	    \centering
	    \begin{subfigure}[]{0.49\linewidth}
			\centering
			\includegraphics[width=\linewidth]{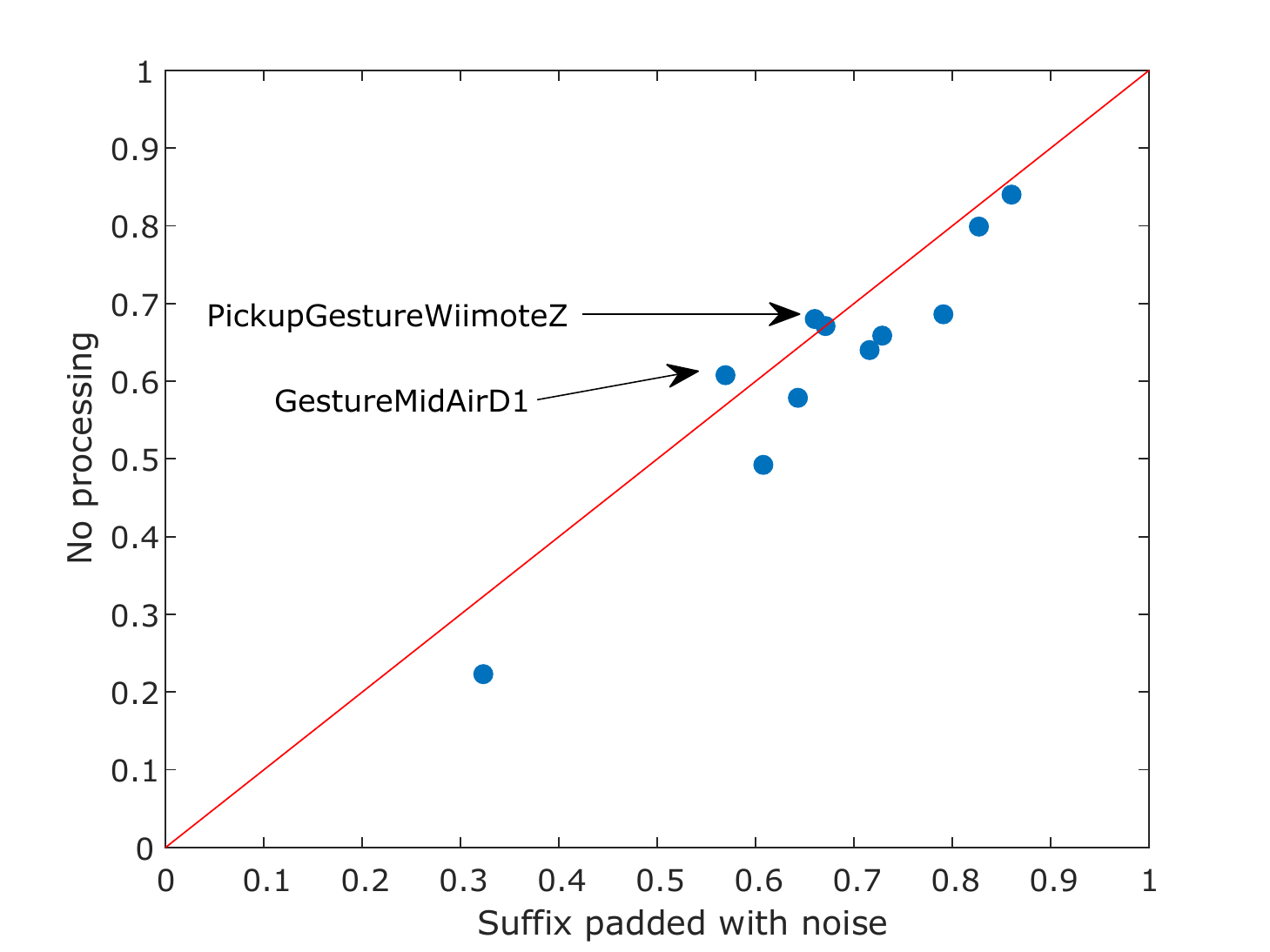}
			\caption{}
			\label{fig:no processing}
		\end{subfigure}
		\hfill
		\begin{subfigure}[]{0.49\linewidth}
			\centering
			\includegraphics[width=\linewidth]{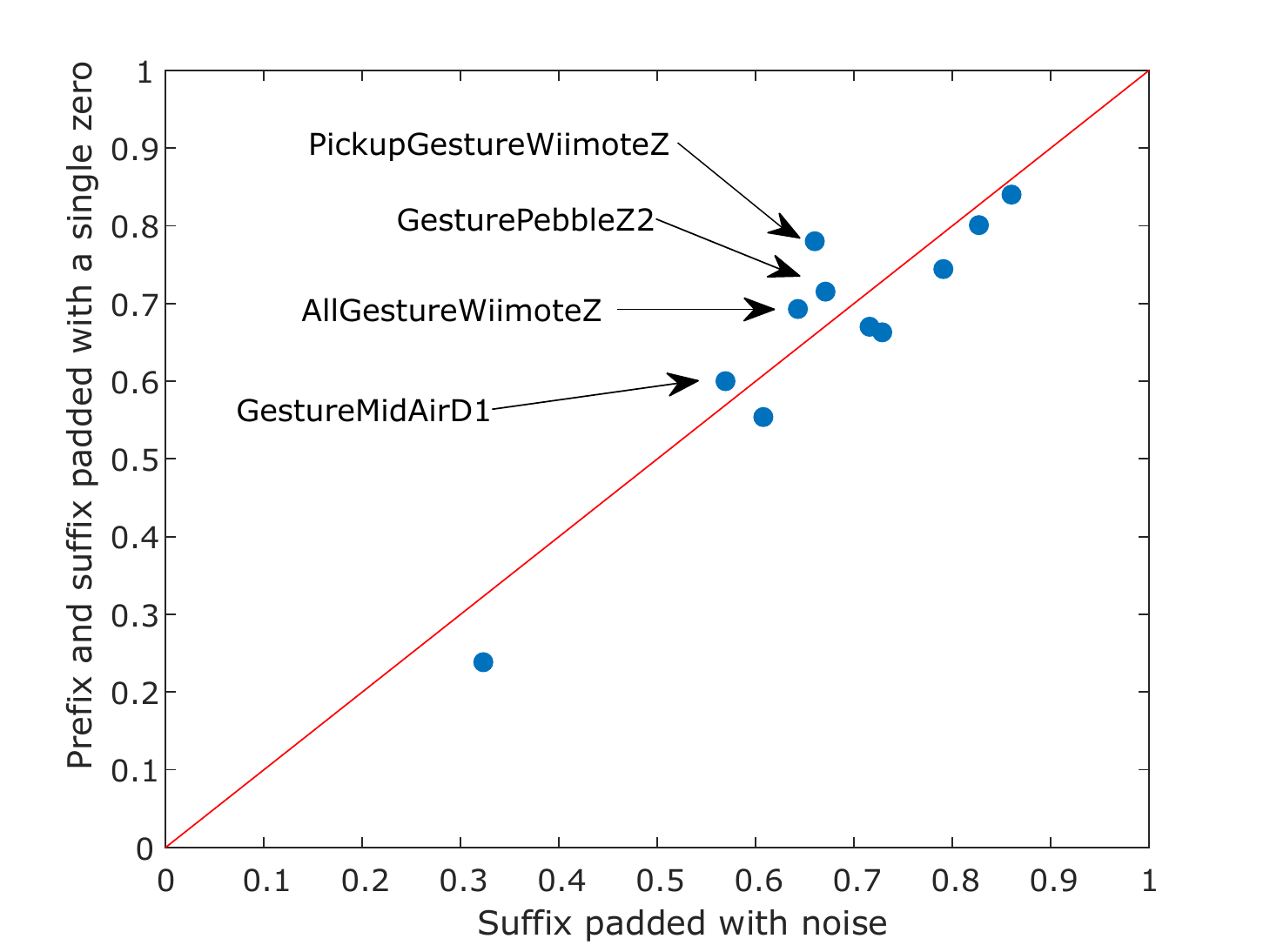}
			\caption{}
			\label{fig:prefix suffix zero}
		\end{subfigure}
		\hfill
		\begin{subfigure}[]{0.49\linewidth}
			\centering
			\includegraphics[width=\linewidth]{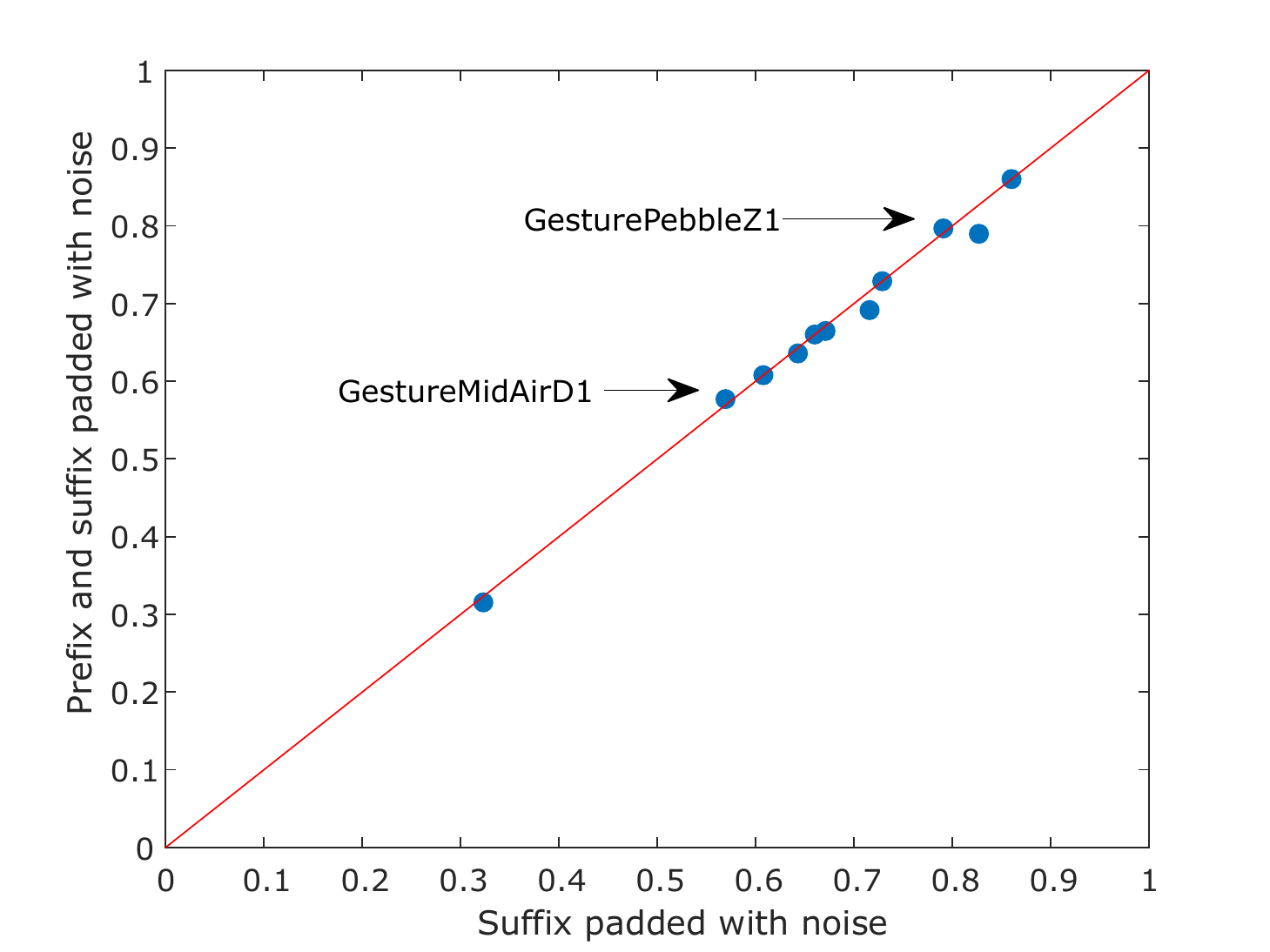}
			\caption{}
			\label{fig:prefix suffix noise}
		\end{subfigure}
		\hfill
		\begin{subfigure}[]{0.49\linewidth}
			\centering
			\includegraphics[width=\linewidth]{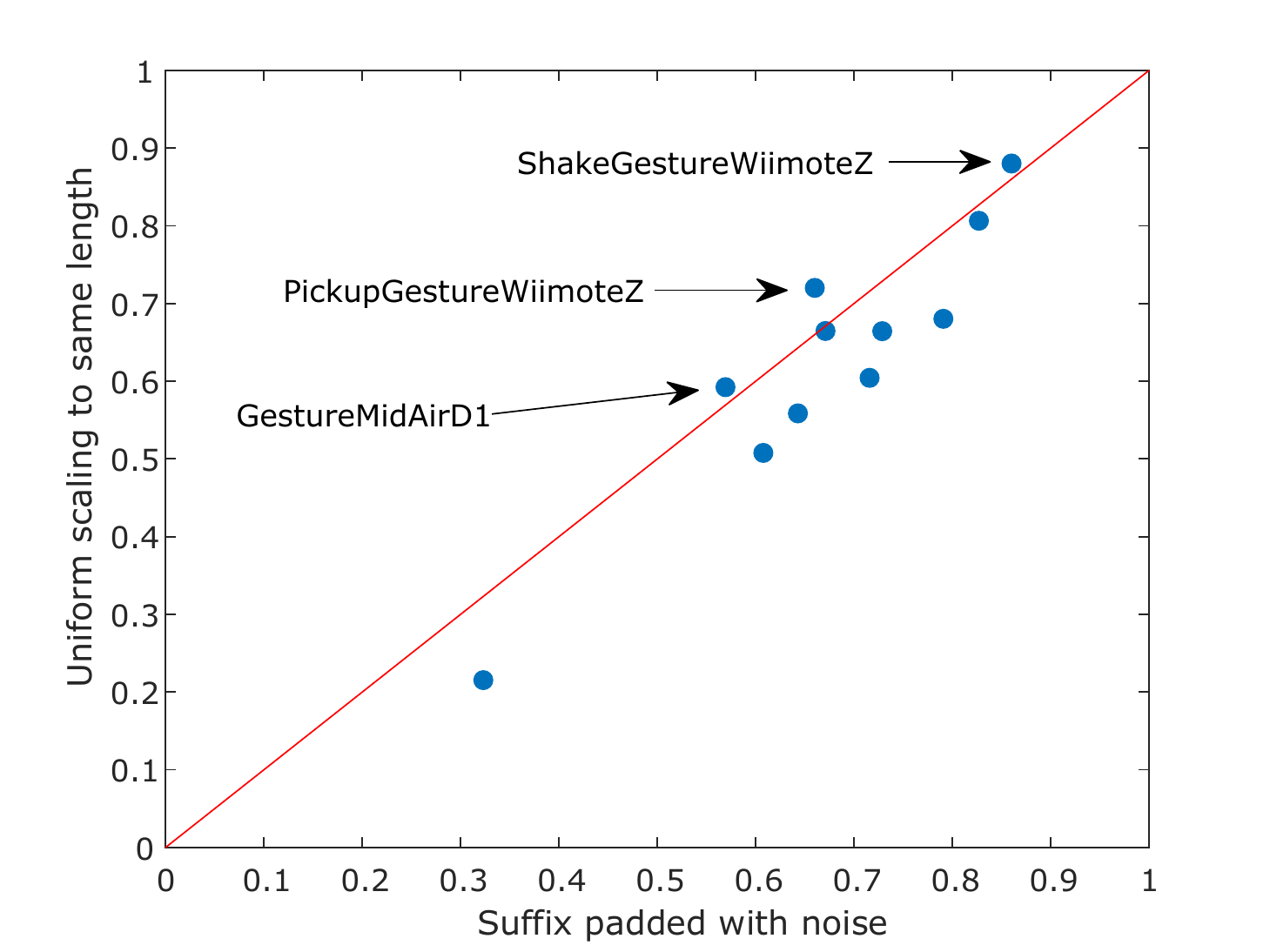}
			\caption{}
			\label{fig:same length rescale}
		\end{subfigure}
	    \caption{Comparison \nndtw{} classification accuracy by padding suffix of the time series with noise with (a) no processing, (b) prefix and suffix padded with a single zero, (c) prefix and suffix padded with noise and (d) uniform scaling the shorter time series to the same length, using the variable lengths time series dataset from the latest UCR time series benchmark archive \citep{UCRArchive2018}}.
	    \label{fig:Comparison of suffix noise}
	\end{figure}

	Figure~\ref{fig:Comparison of suffix noise} plots the classification accuracy of \nndtw{} on the variable length datasets from the latest update to the UCR Archive \citep{UCRArchive2018} using the different processing techniques described in Section \ref{sec:processors}. 
	While it would be dangerous to generalize too far from this small collection of time series from similar domains, 
	these figures show that while padding the suffix with low amplitude noise works best overall, it is outperformed on at least two of the eleven datasets by each of the other approaches. 
	In particular, it does not work well for the \texttt{GestureMidAirD1} dataset, 
	where all the other techniques give better classification accuracy.
	
	\clearpage
	\section{Modified UCR time series datasets}
	\label{app:modified ucr}
\begin{figure}
		\centering
		\begin{subfigure}[]{0.32\linewidth}
			\includegraphics[width=\linewidth]{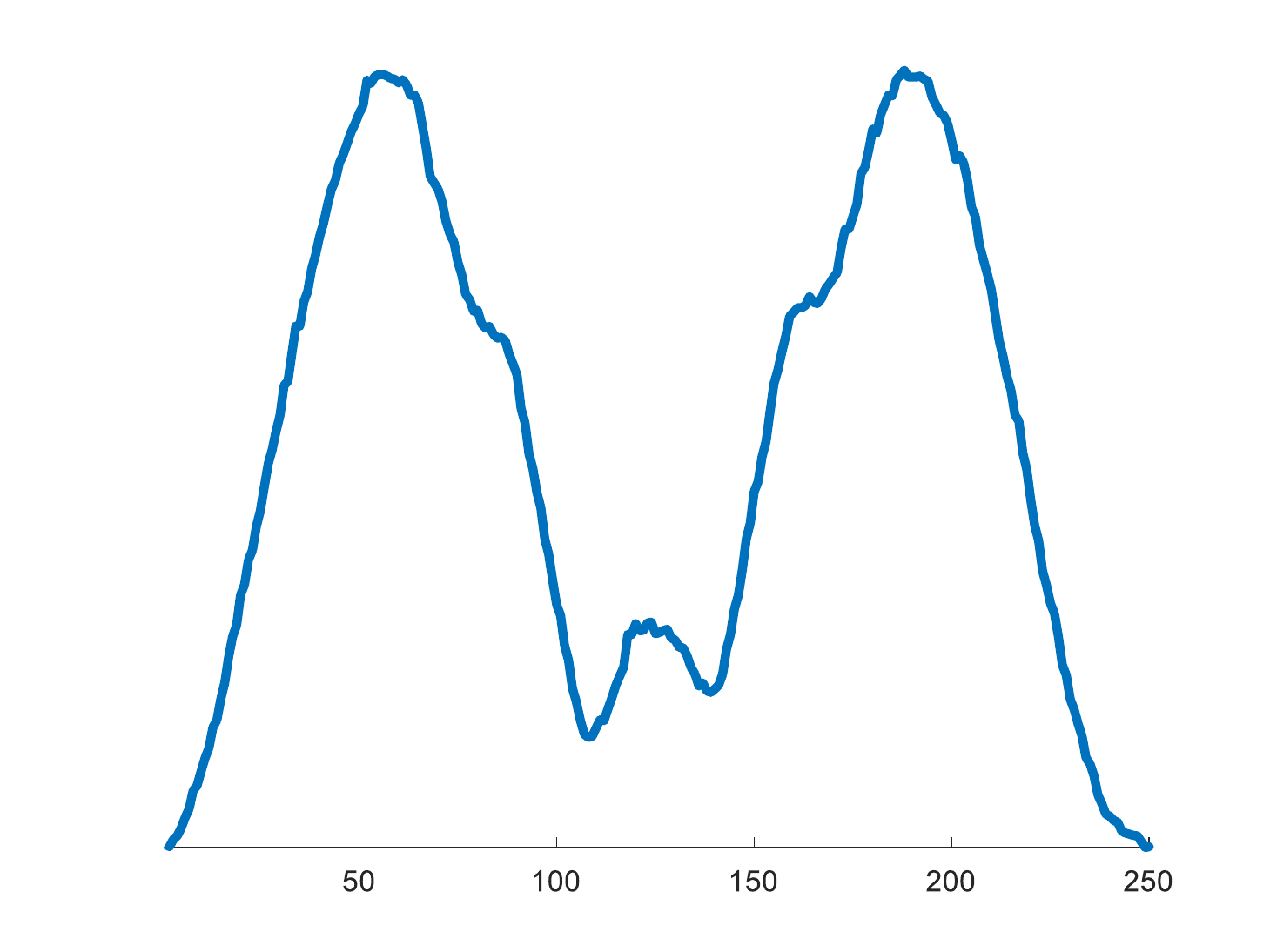}
			\caption{}
			\label{fig:example original}
		\end{subfigure}
		\begin{subfigure}[]{0.32\linewidth}
			\includegraphics[width=\linewidth]{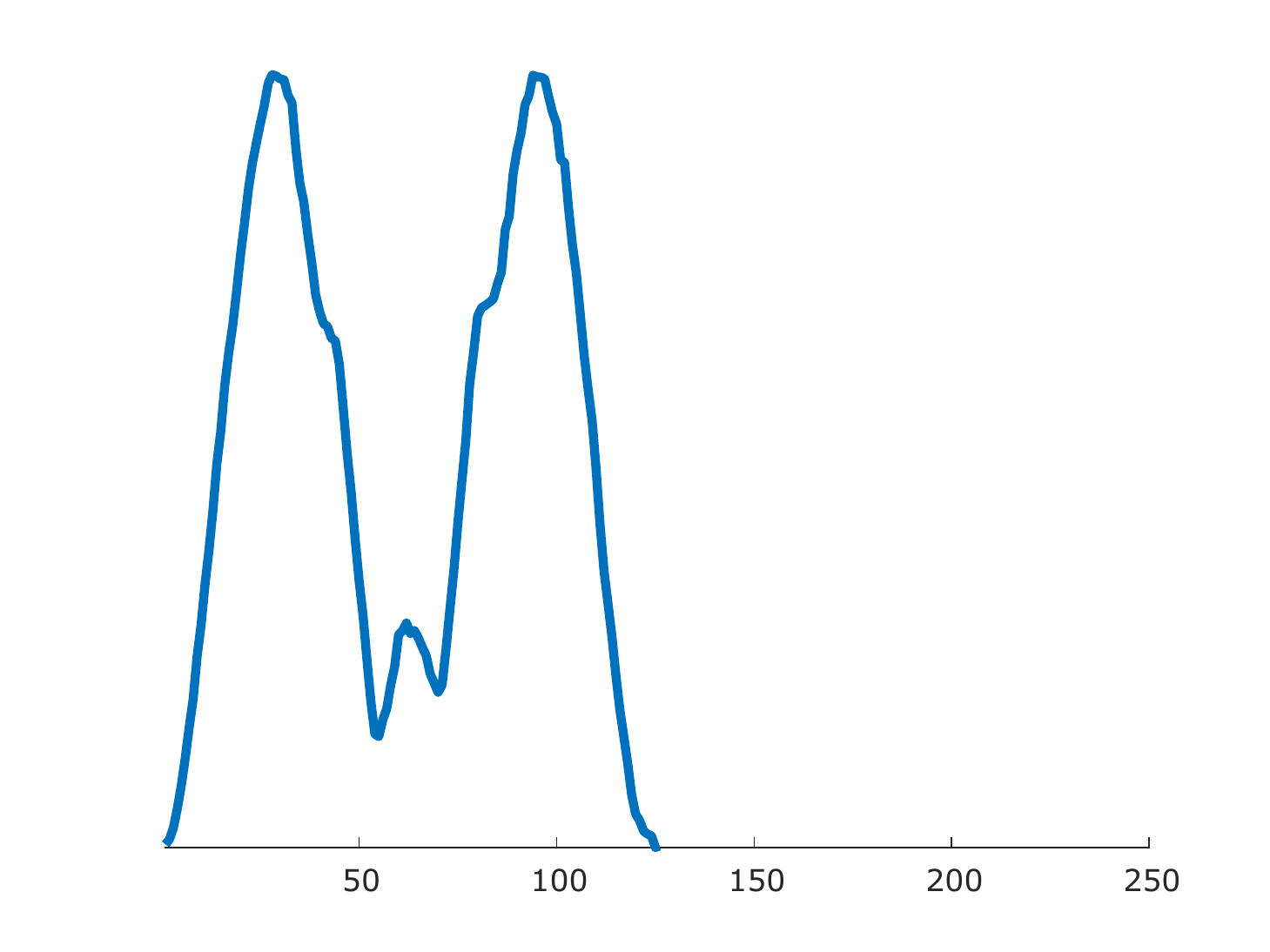}
			\caption{}
			\label{fig:example uniform}
		\end{subfigure}
		\begin{subfigure}[]{0.32\linewidth}
			\includegraphics[width=\linewidth]{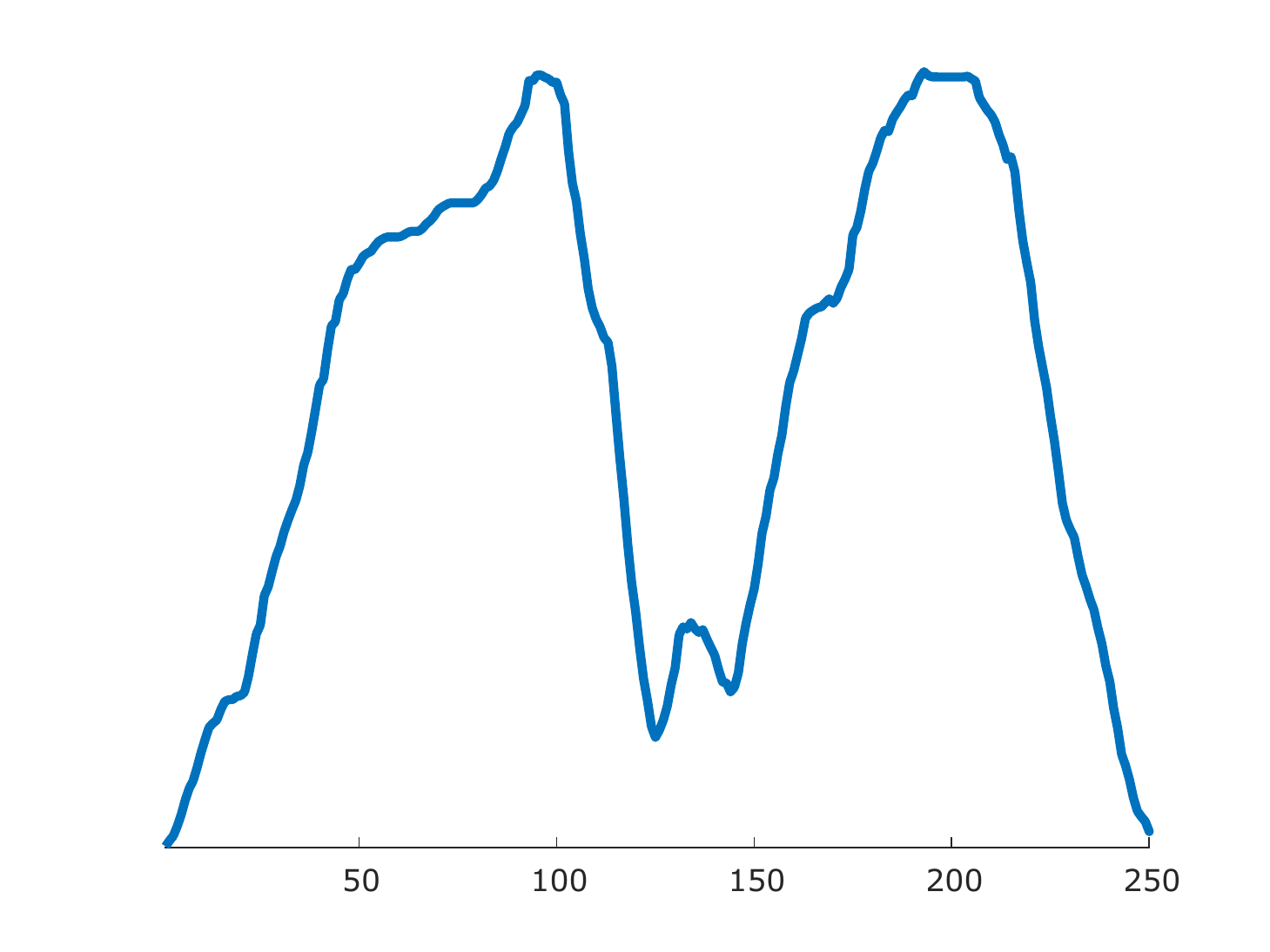}
			\caption{}
			\label{fig:example non uniform}
		\end{subfigure}
		\begin{subfigure}[]{0.32\linewidth}
			\includegraphics[width=\linewidth]{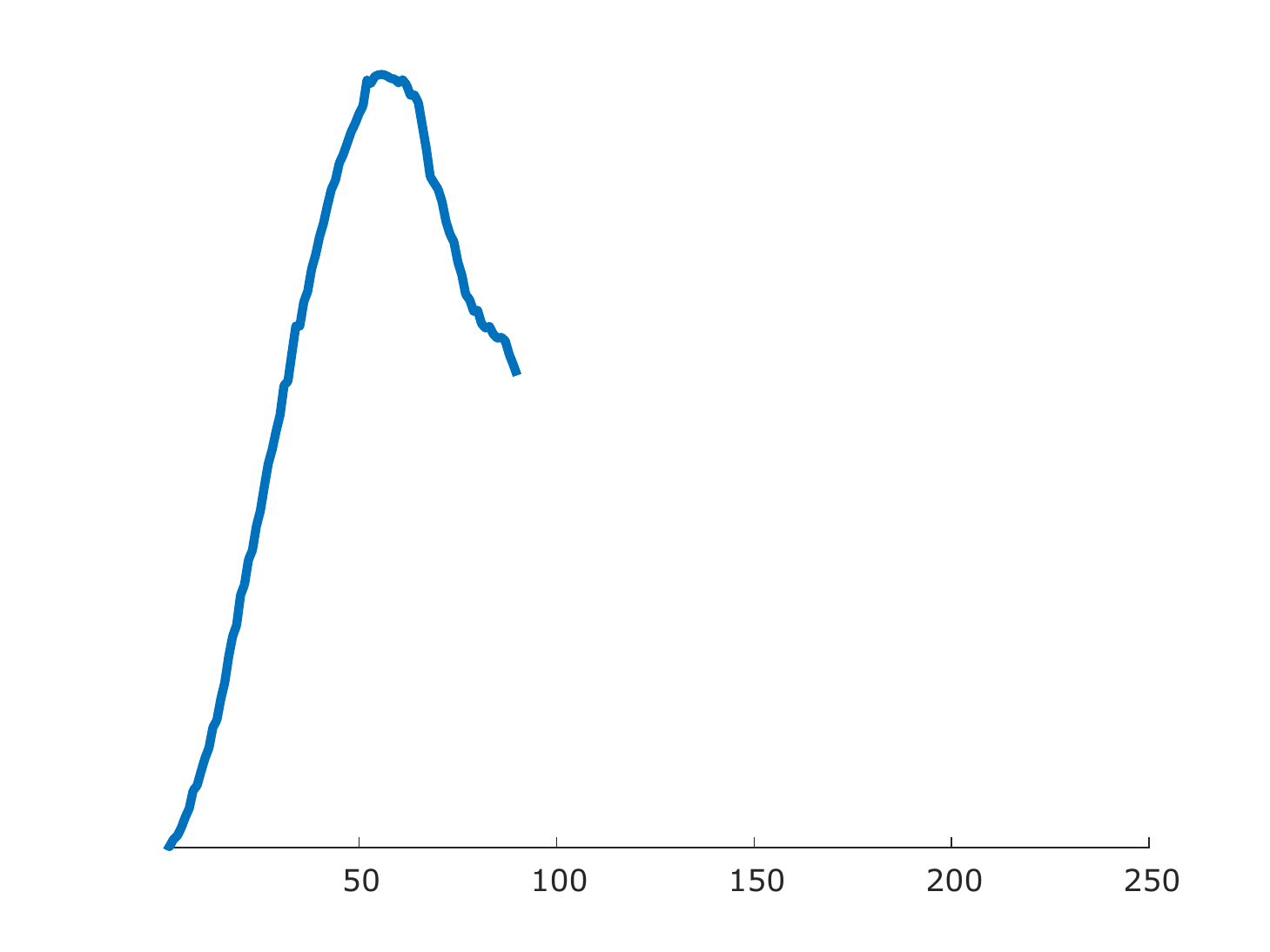}
			\caption{}
			\label{fig:example prefix}
		\end{subfigure}
		\begin{subfigure}[]{0.32\linewidth}
			\includegraphics[width=\linewidth]{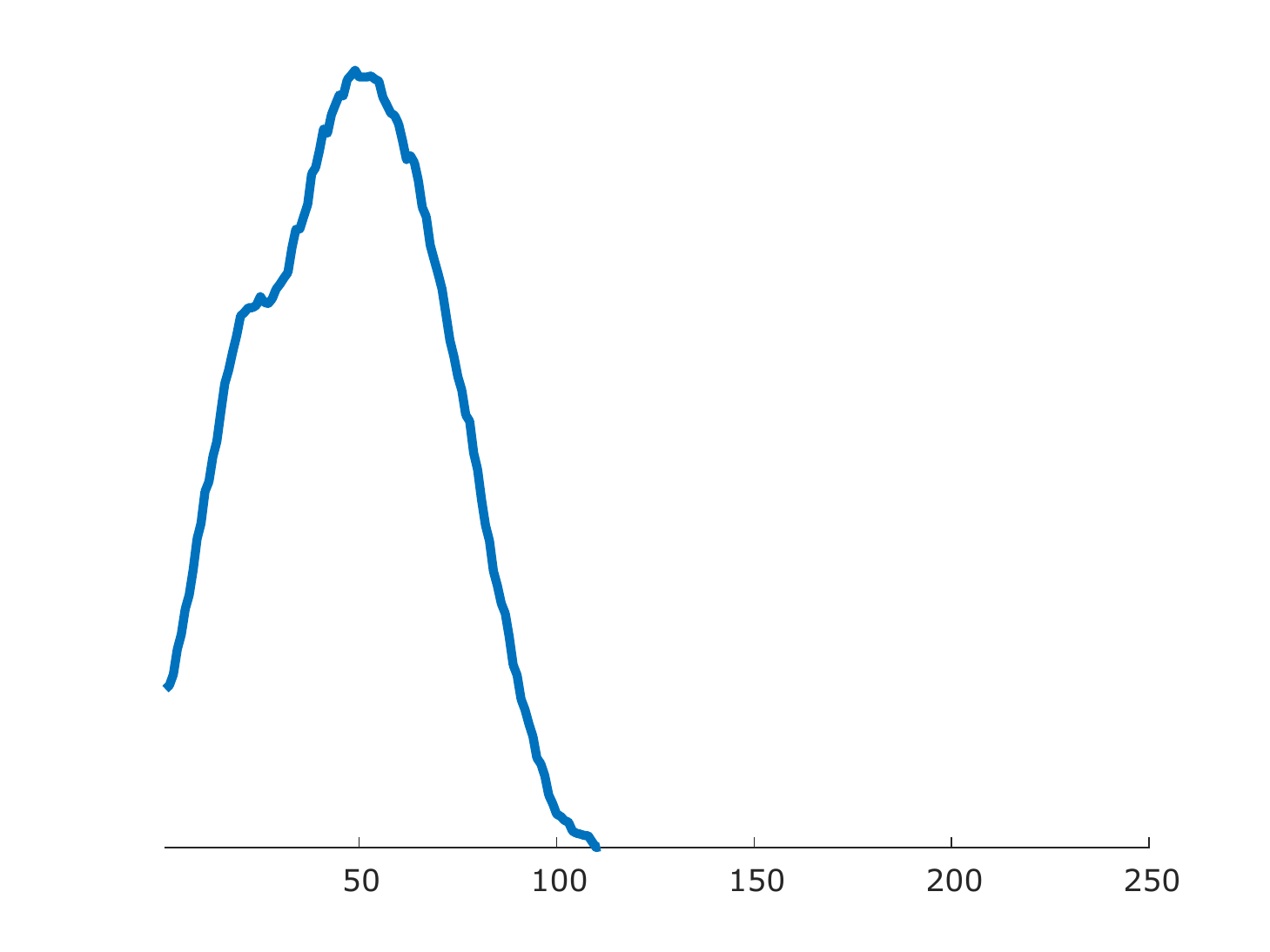}
			\caption{}
			\label{fig:example suffix}
		\end{subfigure}
		\begin{subfigure}[]{0.32\linewidth}
			\includegraphics[width=\linewidth]{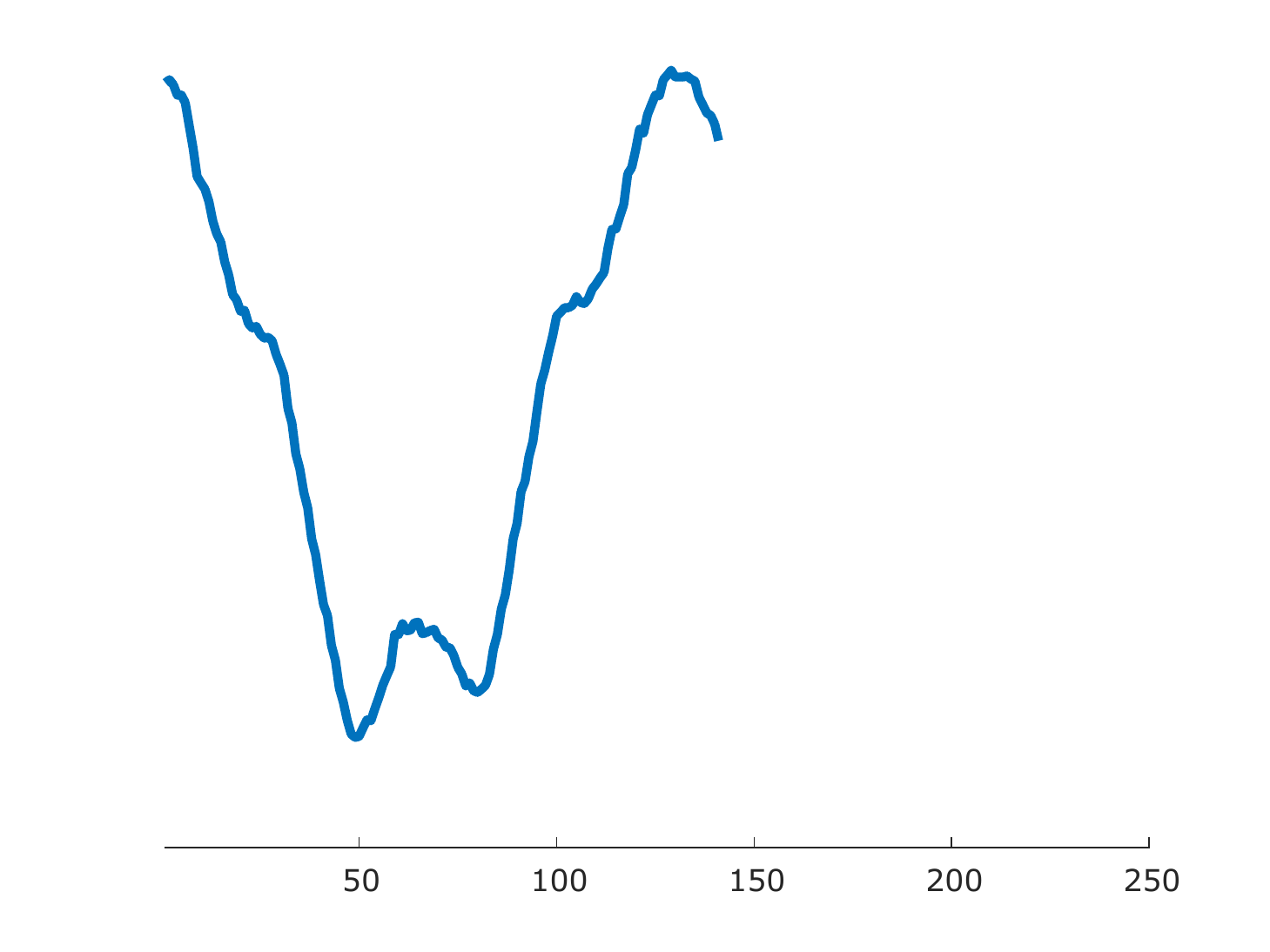}
			\caption{}
			\label{fig:example subsequence}
		\end{subfigure}
		\caption{(a) Original time series from the \texttt{ArrowHead} dataset \citep{ucrarchive}. (b) Uniformly sampled series. (c) Non-uniformly sampled series. (d) Prefix  series. (e) Suffix  series. (f) Subsequence  series.}
		\label{fig:example data}
	\end{figure}

	We modify the 2015 UCR time series archive \citep{ucrarchive} to simulate the different mechanisms.
	This allows us to understand the interactions between each mechanism, preprocessing technique and TSC strategy.
	For each mechanism, for each of the train and test set of each dataset in the archive, 85\% of the instances are randomly selected and modified, leaving 15\% unchanged  
	In this work, we perform five modifications to the datasets:
	
	\begin{enumerate}
		\item \textbf{Uniform Sampling}: We assume that each of the time series data are generated with a fixed frequency.
		Each of the time series is modified by uniform sampling the original time series to a random length, $L'$ sampled from a uniform distribution $U(1,L)$, 
		where $L$ is the length of the time series.
		The points are taken by linearly interpolating at a fixed $L/L'$ interval. 
		An example is illustrated in Figure \ref{fig:example uniform} after modifying the time series in Figure \ref{fig:example original}.
		\item \textbf{Non-Uniform Sampling}: We assume that each of the time series data are generated with varying frequencies.
		Each of the time series is modified by sampling with random varying rates of the original time series.
		The rates are varied using a random walk where the points are taken by linearly interpolating at $t'=t+\delta_{t'}$ intervals, where $t'$ is the next time interval and $\delta_{t'}$ is taken from a normal distribution $\mathcal{N}(\delta_{t}, 0.2)$ with a standard deviation of 0.2 and mean equals the current time step increment $\delta_{t}$, and $\delta_{t=0} = 1$.
		Then the points are taken by linearly interpolating at $\delta_t$ intervals.
		An example is illustrated in Figure \ref{fig:example non uniform} after modifying the time series in Figure \ref{fig:example original}.
		\item \textbf{Prefix}: We assume that there are missing data at the end of the observation.
		Each of the time series is modified by removing a suffix of random length from the original time series.
		The length to be removed is sampled from a uniform distribution $U(1,L-1)$, where $L$ is the length of the time series. 
        An example is illustrated in Figure \ref{fig:example prefix} after modifying the time series in Figure \ref{fig:example original}
		\item \textbf{Suffix}: We assume that there are missing data at the start of the observation.
		Each of the time series is modified by removing prefix of random lengths from the original time series.
		The length to be removed is sampled from a uniform distribution $U(1,L-1)$, where $L$ is the length of the time series.
		An example is illustrated in Figure \ref{fig:example suffix} after modifying the time series in Figure \ref{fig:example original}
		\item \textbf{Subsequence}: We assume that there are missing data at the start and end of the observation.
		Each of the time series is modified by removing prefix and suffix of random lengths from the original time series.
		The length of the subsequence is randomly sampled from a uniform distribution $U(1,L-1)$, where $L$ is the length of the time series.
		Then the length of prefix and suffix to remove is sampled from a uniform distribution $U(1,L')$, where $L'$ is the difference between $L$ and the length of the subsequence.
		An example is illustrated in Figure \ref{fig:example subsequence} after modifying the time series in Figure \ref{fig:example original}.
	\end{enumerate}
	
\end{document}

%% file: preamble.tex
\usepackage[utf8]{inputenc}
\usepackage{amsmath,amssymb,color}
\usepackage[pdftex]{graphicx}
\usepackage{enumitem}
\usepackage{url}
\usepackage{array}
\usepackage[breaklinks=true]{hyperref}
\usepackage{breakcites}
\usepackage{subcaption}
\usepackage[linesnumbered,ruled,algo2e]{algorithm2e}
\usepackage{mathtools}
\usepackage{multirow}
\usepackage{booktabs}
\usepackage{caption}
\usepackage{array}
\usepackage{cite}
\usepackage{natbib}
\usepackage[colorinlistoftodos,prependcaption,textsize=tiny]{todonotes}

\hypersetup{
	colorlinks=false,
	citecolor=black
}

\graphicspath{{images/}}

\newcommand{\nn}{\textsc{NN}}
\newcommand{\nndtw}{\textsc{NN-DTW}}
\newcommand{\nned}{\textsc{NN-ED}}
\newcommand{\nnsbd}{\textsc{NN-SBD}}
\newcommand{\nnssd}{\textsc{NN-SSD}}
\newcommand{\nnus}{\textsc{NN-US}}
\newcommand{\hivecote}{\textsc{HIVE-COTE}}
\newcommand{\cote}{\textsc{COTE}}
\newcommand{\ee}{\textsc{EE}}
\newcommand{\pf}{\textsc{PF}}
\newcommand{\boss}{\textsc{BOSS}}

\newcommand{\dtw}{\textsc{DTW}}
\newcommand{\sbd}{\textsc{SBD}}